\documentclass[fleqn,10pt]{wlscirep}

\usepackage[utf8]{inputenc}
\usepackage{longtable}
\usepackage{array}
\usepackage{tabularx}
\usepackage{url}
\usepackage{textcomp}
\usepackage{placeins}
\usepackage{float}
\usepackage{flafter}
\usepackage{needspace}
\DeclareCaptionFont{npjcaption}{\fontsize{9}{10}\selectfont}
\usepackage{etoolbox}
\usepackage{hyperref}
\usepackage{xurl}

\newcommand{\npjtablefont}{\footnotesize}
\newcommand{\npjtablesetup}{\setlength{\tabcolsep}{3pt}\renewcommand{\arraystretch}{1.12}}
\newcommand{\npjtablehead}[1]{\textbf{#1}}
\newcommand{\ms}[2]{\ensuremath{#1\,{\scriptscriptstyle\pm}\,#2}}
\newcommand{\npjmetric}[2]{\shortstack[c]{#1\\[-1pt]{\scriptsize(#2)}}}

\newcommand{\npjmedian}[2]{\shortstack[c]{#1\\[-1pt]{\scriptsize[#2]}}}
\newcommand{\npjnpct}[2]{\mbox{#1 (#2\%)}}
\newcolumntype{L}[1]{>{\raggedright\arraybackslash}p{#1}}
\newcolumntype{C}[1]{>{\centering\arraybackslash}p{#1}}
\newcolumntype{Y}{>{\centering\arraybackslash}X}

\title{Pulse waveform morphology encodes haemodynamic signatures of impending stroke}

\author[1,2,+]{Cheng Ding}
\author[1,2,+]{Jiaming Liu}
\author[3,4,*]{Jian Wu}
\author[3,4]{Hongxia Xu}
\author[1,2,*]{Daoqiang Zhang}

\affil[1]{College of Artificial Intelligence, Nanjing University of Aeronautics and Astronautics, Nanjing, Jiangsu, China}
\affil[2]{Key Laboratory of Brain-Machine Intelligence Technology, Ministry of Education, Nanjing University of Aeronautics and Astronautics, Nanjing, Jiangsu, China}
\affil[3]{State Key Laboratory of Transvascular Implantation Devices and TIDRI, School of Medicine, Zhejiang University, Hangzhou, 310009, Zhejiang, China}
\affil[4]{Zhejiang Key Laboratory of Medical Imaging Artificial Intelligence, Zhejiang University, Hangzhou, 310058, Zhejiang, China}
\affil[*]{Corresponding authors: wujian2000@zju.edu.cn; dqzhang@nuaa.edu.cn}
\affil[+]{These authors contributed equally to this work.}


\begin{abstract}
Stroke is a disorder of the circulation, yet the haemodynamic state that immediately precedes it has rarely been observed, because continuous physiological recordings before a stroke are seldom available. Photoplethysmography (PPG), recorded by every pulse oximeter and by hundreds of millions of wearable devices, captures the peripheral pulse beat by beat, but its morphology is largely discarded. Here we ask whether pulse waveform morphology encodes haemodynamic signatures of impending stroke. Using in-hospital stroke as a natural experiment that provides continuous PPG before physician-adjudicated stroke recognition, we studied intensive-care (MIMIC-III, $n=176$) and emergency-department (MC-MED, $n=154$) cohorts and described each beat with interpretable descriptors of pulse timing, amplitude, contour and variability. At a nominal 360-min horizon, morphology separated pre-recognition from stable windows with $F_1$ scores of 0.941 and 0.989 (sensitivity 0.998 and 0.980), with models transferred unchanged to the second cohort, and outperformed heart rate alone on identical windows. The signature was dominated by systolic crest time and its change from each patient's own baseline, feature attribution was concordant across cohorts (Spearman $\rho=0.865$), and morphology recorded in the first two hours of care stratified the subsequent clinical course ($F_1$ 0.714 and 0.691). In high-risk patients without stroke, 29\% and 14\% of windows were positive, but five-window persistence rules limited alerts to 17\% and 2\% of patients. The results demonstrate that the routinely recorded pulse thus carries haemodynamic imprint of the period preceding stroke, positioning pulse morphology as a candidate continuous cerebrovascular biomarker.

\end{abstract}

\begin{document}

\raggedbottom
\maketitle
\thispagestyle{empty}

\section*{Introduction}

\begin{figure}[!htbp]
    \centering
    \includegraphics[width=\textwidth]
    {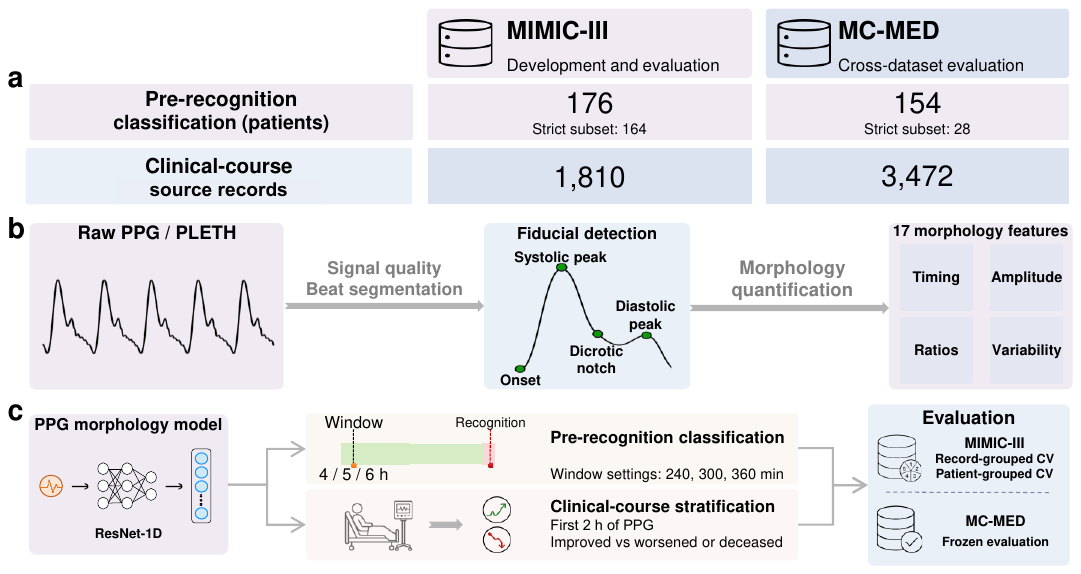}
    \caption{PPG analysis framework.
    \textbf{a}, PPG-linked cohorts and dataset roles. Pre-recognition cohorts are counted as patients and clinical-course cohorts as waveform-linked records. Strict pre-recognition subsets are shown separately.
    \textbf{b}, Waveform processing and extraction of 17 timing, amplitude, ratio and variability features.
    \textbf{c}, ResNet-1D modelling and evaluation. Pre-recognition classification uses nominal 240-, 300- and 360-min window-construction settings relative to documented recognition. The original clinical-course model uses the first 120\,min of PPG after intensive-care admission (MIMIC-III) or emergency-department arrival (MC-MED). Outcomes are improved versus worsened/deceased. The clinical-course counts in panel a describe the waveform-linked source records.}
    \label{fig:study-framework}
\end{figure}

\par Stroke is, at its root, a failure of the circulation to the brain. The brain receives a disproportionate share of cardiac output, has almost no energy reserve and depends on a continuous, pressure-buffered supply. Ischaemic stroke occurs when that supply is interrupted by embolism, thrombosis or hypoperfusion, and haemorrhagic stroke when a vessel wall fails under haemodynamic load. It is therefore unsurprising that the dominant modifiable determinants of stroke are haemodynamic. High systolic blood pressure accounts for more than half of the global disability-adjusted life-years lost to stroke \cite{Feigin2021GBD2019}; visit-to-visit variability and peak systolic pressure predict stroke independently of mean pressure \cite{Rothwell2010BPV}; aortic stiffness predicts fatal stroke beyond conventional risk factors \cite{Laurent2003Stiffness}; and atrial fibrillation couples disordered cardiac rhythm to embolism. Stroke remains the third leading cause of death worldwide, accounting for 7.3 million deaths in 2021 \cite{GBD2021StrokeRiskFactorCollaborators2024}.

\par Yet the haemodynamic view of stroke is almost entirely static. Risk is assigned from clinic blood pressure, from the presence or absence of atrial fibrillation and from scores such as \(\mathrm{CHA}_2\mathrm{DS}_2\text{-}\mathrm{VASc}\), which estimate annual probabilities \cite{Lip2010CHA2DS2VASc,Kim2017CHA2DS2VASc}; models built on structured electronic health records (EHRs) add laboratory values and intermittently charted vital signs \cite{Nwosu2019StrokeEHR,Teoh2018StrokeEHR,Yang2021StrokePrediction}. These tools identify who is vulnerable, but not when. There are good reasons to think that the transition to stroke has a dynamic haemodynamic phase. Transient exposures that provoke sympathetic surges and abrupt pressure changes, such as anger or emotional upset and heavy physical exertion, are associated with stroke onset within the following hour \cite{Smyth2022INTERSTROKE}; hypotension and low cardiac output precipitate ischaemia in vulnerable territories; and new-onset atrial fibrillation, which affects 9--25\% of critically ill patients, creates a substrate for embolism \cite{Noda2023ICUStroke}. Conversely, a cerebral event perturbs the systemic circulation almost immediately. Blood pressure exceeds 140/90\,mmHg in about three-quarters of patients with acute ischaemic stroke and four-fifths of those with intracerebral haemorrhage \cite{Qureshi2008AHR}; rising intracranial pressure produces the Cushing pattern of hypertension with relative bradycardia; and catecholamine surges injure the myocardium and destabilise rhythm, giving rise to the stroke--heart syndrome \cite{Scheitz2022StrokeHeart}. Because cerebral autoregulation is impaired after stroke, the injured brain is in turn exposed to these systemic fluctuations \cite{Intharakham2019Autoregulation}. Haemodynamics are thus both a cause and a consequence of stroke, and the hours surrounding the event should carry their imprint. Whether they do has remained largely unknown for a practical reason: almost no one has a continuous haemodynamic recording of the hours before their stroke.

\par Photoplethysmography (PPG) offers a way to obtain such recordings at scale. Every pulse oximeter records it continuously, as do hundreds of millions of consumer wearables \cite{Charlton2022WearablePPG,Charlton2023WearablePPGRoadmap}, yet bedside monitors use it mainly to derive pulse rate and oxygen saturation. The waveform itself is a compressed record of cardiovascular state. Its upstroke reflects the vigour and duration of left-ventricular ejection and the compliance of the arterial tree; the timing of its peak and the shape of its contour reflect peripheral vascular tone and the arrival of reflected waves; and its beat-to-beat amplitude reflects rhythm, respiratory--circulatory interaction and autonomic modulation \cite{Allen2007PPG,Charlton2022Hemodynamics}. Indices derived from the waveform and its derivatives track vasoactive drug effects and vascular ageing \cite{Takazawa1998APG}, and waveform analysis has been used to estimate blood pressure, to infer cardiac electrical activity and to detect atrial fibrillation at population scale \cite{Elgendi2024PPGBloodPressure,Ding2026CardioPPG,Perez2019AppleHeart,Guo2019HuaweiHeart}. If the period preceding stroke has a haemodynamic signature, the pulse contour is where it should appear, and the signal is already being recorded.

\par In-hospital stroke provides a natural experiment in which this hypothesis can be tested. Between roughly 2\% and 7\% of strokes begin in patients who are already in hospital \cite{Cumbler2014InHospital,Akbik2020InHospital,Noda2023ICUStroke}, many of whom are continuously monitored for other reasons. For these patients, pulse oximetry has recorded the hours before the event, and earlier recordings from the same monitoring episode provide each patient's own reference state. The design has one defining constraint. Physiological onset is rarely known in hospitalised patients, whose deficits are often masked by sedation, delirium or encephalopathy \cite{Noda2023ICUStroke}, and the interval from last seen well to recognition is longer than for strokes in the community \cite{Correia2025DELAY}. We therefore anchored every analysis to a physician-adjudicated time of documented stroke recognition. The period before this anchor contains whatever haemodynamic changes precede and accompany the event, and the question becomes whether the pulse waveform in that period differs systematically from the same patient's earlier state.

\par Here we ask whether pulse waveform morphology encodes haemodynamic signatures of impending stroke (Fig.~\ref{fig:study-framework}). We studied two waveform-linked clinical resources representing different points of entry into acute care, intensive care (MIMIC-III) \cite{Johnson2016MIMICIII,Moody2020MIMICIIIWaveform} and the emergency department (MC-MED) \cite{Kansal2025MCMED,Kansal2025MCMEDDataset}, and described each pulse with 17 interpretable descriptors of timing, amplitude, contour and beat-to-beat variability, extending our earlier analysis \cite{Liu2026InHospitalStroke}. \textbf{Throughout, stroke time refers to documented recognition in the clinical record, not physiological onset.} We developed models in MIMIC-III and applied them without modification to MC-MED; asked whether the information resides in pulse shape rather than heart rate (HR); asked whether the same morphology, recorded after admission, stratifies the subsequent clinical course; identified the haemodynamic features that carry these signals and tested their agreement across cohorts; and used simultaneous electrocardiography (ECG) to separate the dominant feature, systolic crest time, from heart rate. Finally, we examined the specificity of the signature under strict patient-grouped evaluation and in high-risk patients without stroke, and related it to patient-level alerts and to the monitoring actually available. Together, these analyses show that the routinely discarded pulse carries a reproducible, physiologically interpretable imprint of the period preceding stroke, expressed chiefly as change within each patient.

\FloatBarrier

\section*{Results}

\Needspace{5\baselineskip}
\subsection*{In-hospital stroke as a natural experiment for pre-stroke haemodynamics}

\par We drew on two de-identified resources that link routinely collected clinical data to continuous bedside monitoring (Fig.~\ref{fig:study-framework}a). The MIMIC-III Clinical Database and its Waveform Database Matched Subset cover critical care at Beth Israel Deaconess Medical Center between 2001 and 2012, with 22,317 waveform records from 10,282 patients \cite{Johnson2016MIMICIII,Moody2020MIMICIIIWaveform}; MC-MED covers 118,385 monitored adult emergency-department visits by 70,545 patients at Stanford Health Care between 2020 and 2022 \cite{Kansal2025MCMED,Kansal2025MCMEDDataset}. PPG was the only waveform used. Models were developed in MIMIC-III and applied to MC-MED without retraining, so that the second cohort tests whether what is learned in one care setting holds in another.

\par Within these resources we identified patients whose stroke was recognised while they were being monitored. Candidate recognition times were extracted from clinical notes by a large language model (LLM)-assisted procedure and adjudicated by physicians, with independent review of a stratified sample by two neurologists (Methods). Of 1,934 candidates in MIMIC-III and 1,119 in MC-MED, adjudication confirmed 334 and 325, of whom 176 and 154 had temporally aligned PPG of adequate quality before the recognition anchor; these patients form the pre-recognition (warning) cohorts. A second set of waveform-linked records, 1,810 in MIMIC-III and 3,472 in MC-MED, captured PPG early after admission and was used to study the subsequent clinical course.

\par The two cohorts sample distinct haemodynamic contexts (Table~\ref{tab:cohort-characteristics}). Every MIMIC-III warning patient carried an admission-level haemorrhagic diagnostic code and three-quarters had recorded hypertension, whereas the MC-MED warning cohort was older and predominantly ischaemic. Patients retained after waveform screening resembled those excluded in age and mortality, with a few differences: retained MIMIC-III patients more often had recorded hypertension (76.7\% versus 66.5\%, \(P=0.039\)), and retained MC-MED encounters were longer (7.3 versus 6.0\,h, \(P=0.008\)) and less often had recorded atrial fibrillation (9.7\% versus 18.1\%, \(P=0.038\)).

\begin{table}[H]
\caption{Demographic and clinical characteristics of the PPG-linked study cohorts.}
\label{tab:cohort-characteristics}
\centering
\npjtablefont
\setlength{\tabcolsep}{2.5pt}
\renewcommand{\arraystretch}{1.10}

\begin{tabularx}{\textwidth}{@{}
    L{4.35cm}
    Y
    Y
    Y
    Y
@{}}
\toprule
\npjtablehead{Characteristic} &
\multicolumn{2}{c}{\npjtablehead{Warning cohort}} &
\multicolumn{2}{c}{\npjtablehead{Clinical-course cohort}} \\
\cmidrule(lr){2-3}\cmidrule(lr){4-5}
&
\npjtablehead{MIMIC-III} &
\npjtablehead{MC-MED} &
\npjtablehead{MIMIC-III} &
\npjtablehead{MC-MED} \\
\midrule

\multicolumn{5}{@{}l}{\textbf{Cohort size}} \\[-1pt]
Eligible cohort, \(n\)
& 176
& 154
& 1,810
& 3,472 \\

\addlinespace[3pt]
\multicolumn{5}{@{}l}{\textbf{Age}} \\[-1pt]
Mean, years
& \npjmetric{66.9}{15.3}
& \npjmetric{71.0}{15.0}
& \npjmetric{67.1}{14.2}
& \npjmetric{71.0}{16.8} \\
Median, years
& \npjmedian{68.8}{54.2--79.6}
& \npjmedian{75.0}{61.0--83.0}
& \npjmedian{68.0}{56.0--79.0}
& \npjmedian{75.0}{62.0--85.0} \\

\addlinespace[1pt]
\multicolumn{5}{@{}l}{\textit{Binary age grouping}} \\[-1pt]
\(<65\) years, \(n\) (\%)
& \npjnpct{76}{43.2}
& \npjnpct{47}{30.5}
& \npjnpct{762}{42.1}
& \npjnpct{994}{28.6} \\
\(\geq65\) years, \(n\) (\%)
& \npjnpct{100}{56.8}
& \npjnpct{107}{69.5}
& \npjnpct{1,048}{57.9}
& \npjnpct{2,478}{71.4} \\

\addlinespace[1pt]
\multicolumn{5}{@{}l}{\textit{Four-category age grouping}} \\[-1pt]
18--39 years, \(n\) (\%)
& \npjnpct{9}{5.1}
& \npjnpct{4}{2.6}
& \npjnpct{64}{3.5}
& \npjnpct{215}{6.2} \\
40--59 years, \(n\) (\%)
& \npjnpct{52}{29.5}
& \npjnpct{30}{19.5}
& \npjnpct{499}{27.6}
& \npjnpct{534}{15.4} \\
60--79 years, \(n\) (\%)
& \npjnpct{75}{42.6}
& \npjnpct{70}{45.5}
& \npjnpct{809}{44.7}
& \npjnpct{1,473}{42.4} \\
\(\geq80\) years, \(n\) (\%)
& \npjnpct{40}{22.7}
& \npjnpct{50}{32.5}
& \npjnpct{438}{24.2}
& \npjnpct{1,250}{36.0} \\

\addlinespace[3pt]
\midrule
\addlinespace[1pt]
\multicolumn{5}{@{}l}{\textbf{Sex}} \\[-1pt]
Female, \(n\) (\%)
& \npjnpct{88}{50.0}
& \npjnpct{74}{48.1}
& \npjnpct{894}{49.4}
& \npjnpct{1,612}{46.4} \\
Male, \(n\) (\%)
& \npjnpct{88}{50.0}
& \npjnpct{80}{51.9}
& \npjnpct{916}{50.6}
& \npjnpct{1,850}{53.3} \\
Unknown, \(n\) (\%)
& \npjnpct{0}{0.0}
& \npjnpct{0}{0.0}
& \npjnpct{0}{0.0}
& \npjnpct{10}{0.3} \\

\addlinespace[3pt]
\midrule
\addlinespace[1pt]
\multicolumn{5}{@{}l}{\textbf{Race}} \\[-1pt]
White, \(n\) (\%)
& \npjnpct{124}{70.5}
& \npjnpct{74}{48.1}
& \npjnpct{1,262}{69.7}
& \npjnpct{1,818}{52.4} \\
Asian, \(n\) (\%)
& \npjnpct{5}{2.8}
& \npjnpct{43}{27.9}
& \npjnpct{53}{2.9}
& \npjnpct{726}{20.9} \\
Black or African American, \(n\) (\%)
& \npjnpct{13}{7.4}
& \npjnpct{7}{4.5}
& \npjnpct{147}{8.1}
& \npjnpct{177}{5.1} \\
Pacific Islander, \(n\) (\%)
& \npjnpct{0}{0.0}
& \npjnpct{2}{1.3}
& \npjnpct{0}{0.0}
& \npjnpct{33}{1.0} \\
American Indian/Alaska Native, \(n\) (\%)
& \npjnpct{0}{0.0}
& \npjnpct{0}{0.0}
& \npjnpct{0}{0.0}
& \npjnpct{10}{0.3} \\
Other, \(n\) (\%)
& \npjnpct{21}{11.9}
& \npjnpct{28}{18.2}
& \npjnpct{204}{11.3}
& \npjnpct{674}{19.4} \\
Declines to state, \(n\) (\%)
& \npjnpct{0}{0.0}
& \npjnpct{0}{0.0}
& \npjnpct{9}{0.5}
& \npjnpct{5}{0.1} \\
Unknown, \(n\) (\%)
& \npjnpct{13}{7.4}
& \npjnpct{0}{0.0}
& \npjnpct{135}{7.5}
& \npjnpct{29}{0.8} \\

\addlinespace[3pt]
\midrule
\addlinespace[1pt]
\multicolumn{5}{@{}l}{\textbf{Comorbidities}} \\[-1pt]
Hypertension, \(n\) (\%)
& \npjnpct{135}{76.7}
& \npjnpct{57}{37.0}
& \npjnpct{1,030}{56.9}
& \npjnpct{1,254}{36.1} \\
Diabetes, \(n\) (\%)
& \npjnpct{39}{22.2}
& \npjnpct{21}{13.6}
& \npjnpct{339}{18.7}
& \npjnpct{562}{16.2} \\
Hyperlipidaemia, \(n\) (\%)
& \npjnpct{70}{39.8}
& \npjnpct{41}{26.6}
& \npjnpct{608}{33.6}
& \npjnpct{721}{20.8} \\
Chronic kidney disease, \(n\) (\%)
& \npjnpct{14}{8.0}
& \npjnpct{14}{9.1}
& \npjnpct{177}{9.8}
& \npjnpct{177}{5.1} \\
Ischaemic heart disease, \(n\) (\%)
& \npjnpct{30}{17.0}
& \npjnpct{26}{16.9}
& \npjnpct{410}{22.7}
& \npjnpct{403}{11.6} \\

\addlinespace[3pt]
\midrule
\addlinespace[1pt]
\multicolumn{5}{@{}l}{\textbf{Diagnostic code groups, \(n\) (\%)}} \\[-1pt]
Ischaemic/occlusive family only & \npjnpct{0}{0.0} & \npjnpct{105}{68.2} & \npjnpct{456}{25.2} & \npjnpct{1,109}{31.9} \\
Haemorrhagic family only & \npjnpct{135}{76.7} & \npjnpct{42}{27.3} & \npjnpct{1,083}{59.8} & \npjnpct{1,244}{35.8} \\
Both coding families & \npjnpct{41}{23.3} & \npjnpct{0}{0.0} & \npjnpct{271}{15.0} & \npjnpct{0}{0.0} \\
Other recorded diagnoses & \npjnpct{0}{0.0} & \npjnpct{7}{4.5} & \npjnpct{0}{0.0} & \npjnpct{1,119}{32.2} \\
\bottomrule
\end{tabularx}
\end{table}

\subsection*{Pulse morphology distinguishes the pre-recognition haemodynamic state}

\par To ask whether the pulse changes before stroke recognition, we compared each patient's PPG in the hours before the recognition anchor with PPG recorded earlier in the same monitoring episode. Windows within a nominal horizon of 240, 300 or 360\,min before the anchor were labelled pre-recognition (warning) and earlier windows, up to 480\,min before the anchor, were labelled stable; a transition buffer at the class boundary and a 15-min blind zone before the anchor were excluded (Methods). Because longer horizons enlarge the warning interval at the expense of the stable one, warning windows formed the majority of labelled windows, increasingly so as the horizon lengthened.

\par Pulse morphology separated the two states in both cohorts (Table~\ref{tab:warning-performance}). In MIMIC-III, separation strengthened with the horizon, $F_1$ rising from 0.80 at 240\,min to 0.94 at 360\,min. Applied without retraining to MC-MED, the same models reached $F_1$ values between 0.93 and 0.99, and at 360\,min sensitivity was 0.998 in MIMIC-III and 0.980 in MC-MED. That a model learned in intensive care transferred to an emergency-department population, recorded a decade later and dominated by a different stroke subtype, suggests that the morphological pattern it captured is not confined to one institution or setting.

\par Morphology outperformed the available-component \(\mathrm{CHA}_2\mathrm{DS}_2\text{-}\mathrm{VASc}\) score and three structured-EHR models at every horizon in both cohorts (Table~\ref{tab:warning-performance}), but the comparison is more informative for what it reveals about the references than for the margin. A risk score is constant across a monitoring episode and cannot, by construction, distinguish earlier from later windows in the same patient. The EHR models, which add intermittently charted vital signs and laboratory values, achieved high $F_1$ at longer horizons largely through near-complete recall and proved unstable under transfer, one of them collapsing to an $F_1$ of 0.13 in MC-MED at 360\,min. The pulse waveform, by contrast, changes from beat to beat within the episode, which is what a time-resolved signature requires.

\begingroup\npjtablefont\npjtablesetup\renewcommand{\arraystretch}{1.06}
\setlength{\LTpre}{6pt}\setlength{\LTpost}{6pt}
\begin{longtable}{@{}L{\dimexpr0.4\textwidth-4.5pt\relax}*{3}{C{\dimexpr0.2\textwidth-4.5pt\relax}}@{}}
\caption{Performance of PPG and EHR reference models for stroke warning across three temporal settings.}\label{tab:warning-performance}\\

\toprule
\textbf{Model} & \textbf{$F_1$} & \textbf{Precision} & \textbf{Recall} \\
\midrule
\endfirsthead
\multicolumn{4}{@{}l}{Table~\thetable (continued)}\\

\toprule
\textbf{Model} & \textbf{$F_1$} & \textbf{Precision} & \textbf{Recall} \\
\midrule
\endhead

\multicolumn{4}{@{}l}{\textbf{a, Nominal 240-min setting}} \\*

\multicolumn{4}{@{}l}{\textit{MIMIC-III (internal validation)}} \\*[-1pt]

\textbf{PPG morphology}
& \ms{0.7956}{0.0027}
& \ms{0.6681}{0.0041}
& \ms{0.9833}{0.0106} \\

\(\mathrm{CHA}_2\mathrm{DS}_2\text{-}\mathrm{VASc}\)
& \ms{0.6570}{0.0250}
& \ms{0.5890}{0.0204}
& \ms{0.7440}{0.0388} \\

Nwosu--EHR
& \ms{0.6650}{0.0314}
& \ms{0.5570}{0.0378}
& \ms{0.8240}{0.0344} \\

Teoh--EHR
& \ms{0.7260}{0.0268}
& \ms{0.5730}{0.0316}
& \ms{0.9920}{0.0066} \\

Yang--EHR
& \ms{0.7250}{0.0268}
& \ms{0.5710}{0.0337}
& \ms{0.9920}{0.0066} \\

\addlinespace[1pt]
\multicolumn{4}{@{}l}{\textit{MC-MED (frozen cross-dataset evaluation)}} \\*[-1pt]

\textbf{PPG morphology}
& \ms{0.9256}{0.0211}
& \ms{0.9179}{0.0020}
& \ms{0.9341}{0.0413} \\

\(\mathrm{CHA}_2\mathrm{DS}_2\text{-}\mathrm{VASc}\)
& \ms{0.8030}{0.0321}
& \ms{0.8640}{0.0191}
& \ms{0.7500}{0.0503} \\

Nwosu--EHR
& \ms{0.7940}{0.0385}
& \ms{0.9000}{0.0416}
& \ms{0.7110}{0.0495} \\

Teoh--EHR
& \ms{0.7440}{0.0441}
& \ms{0.9060}{0.0390}
& \ms{0.6320}{0.0559} \\

Yang--EHR
& \ms{0.8380}{0.0339}
& \ms{0.8610}{0.0416}
& \ms{0.8160}{0.0464} \\
\addlinespace[3pt]
\midrule
\multicolumn{4}{@{}l}{\textbf{b, Nominal 300-min setting}} \\*

\multicolumn{4}{@{}l}{\textit{MIMIC-III (internal validation)}} \\*[-1pt]

\textbf{PPG morphology}
& \ms{0.8759}{0.0105}
& \ms{0.8028}{0.0125}
& \ms{0.9647}{0.0360} \\

\(\mathrm{CHA}_2\mathrm{DS}_2\text{-}\mathrm{VASc}\)
& \ms{0.7160}{0.0235}
& \ms{0.6910}{0.0196}
& \ms{0.7420}{0.0372} \\

Nwosu--EHR
& \ms{0.8080}{0.0230}
& \ms{0.6810}{0.0298}
& \ms{0.9930}{0.0054} \\

Teoh--EHR
& \ms{0.8060}{0.0212}
& \ms{0.6780}{0.0273}
& \ms{0.9930}{0.0054} \\

Yang--EHR
& \ms{0.8060}{0.0214}
& \ms{0.6790}{0.0304}
& \ms{0.9930}{0.0054} \\

\addlinespace[1pt]
\multicolumn{4}{@{}l}{\textit{MC-MED (frozen cross-dataset evaluation)}} \\*[-1pt]

\textbf{PPG morphology}
& \ms{0.9595}{0.0151}
& \ms{0.9802}{0.0023}
& \ms{0.9401}{0.0294} \\

\(\mathrm{CHA}_2\mathrm{DS}_2\text{-}\mathrm{VASc}\)
& \ms{0.8410}{0.0311}
& \ms{0.9380}{0.0161}
& \ms{0.7620}{0.0446} \\

Nwosu--EHR
& \ms{0.8690}{0.0316}
& \ms{0.9690}{0.0209}
& \ms{0.7870}{0.0467} \\

Teoh--EHR
& \ms{0.9340}{0.0288}
& \ms{0.9070}{0.0242}
& \ms{0.9630}{0.0337} \\

Yang--EHR
& \ms{0.9470}{0.0260}
& \ms{0.9210}{0.0247}
& \ms{0.9740}{0.0270} \\
\addlinespace[3pt]
\midrule
\multicolumn{4}{@{}l}{\textbf{c, Nominal 360-min setting}} \\*

\multicolumn{4}{@{}l}{\textit{MIMIC-III (internal validation)}} \\*[-1pt]

\textbf{PPG morphology}
& \ms{0.9406}{0.0015}
& \ms{0.8894}{0.0013}
& \ms{0.9981}{0.0025} \\

\(\mathrm{CHA}_2\mathrm{DS}_2\text{-}\mathrm{VASc}\)
& \ms{0.7650}{0.0222}
& \ms{0.7950}{0.0163}
& \ms{0.7380}{0.0334} \\

Nwosu--EHR
& \ms{0.8840}{0.0186}
& \ms{0.7980}{0.0265}
& \ms{0.9920}{0.0059} \\

Teoh--EHR
& \ms{0.8800}{0.0196}
& \ms{0.7930}{0.0278}
& \ms{0.9890}{0.0064} \\

Yang--EHR
& \ms{0.8880}{0.0189}
& \ms{0.8010}{0.0278}
& \ms{0.9950}{0.0048} \\

\addlinespace[1pt]
\multicolumn{4}{@{}l}{\textit{MC-MED (frozen cross-dataset evaluation)}} \\*[-1pt]

\textbf{PPG morphology}
& \ms{0.9888}{0.0025}
& \ms{0.9975}{0.0008}
& \ms{0.9804}{0.0054} \\

\(\mathrm{CHA}_2\mathrm{DS}_2\text{-}\mathrm{VASc}\)
& \ms{0.8550}{0.0291}
& \ms{0.9850}{0.0008}
& \ms{0.7560}{0.0444} \\

Nwosu--EHR
& \ms{0.9820}{0.0107}
& \ms{0.9880}{0.0041}
& \ms{0.9770}{0.0145} \\

Teoh--EHR
& \ms{0.1300}{0.0464}
& \ms{1.0000}{0.0000}
& \ms{0.0700}{0.0268} \\

Yang--EHR
& \ms{0.8160}{0.0352}
& \ms{0.9840}{0.0133}
& \ms{0.6980}{0.0492} \\
\bottomrule

\end{longtable}
\endgroup

\par
\begin{figure}[H]
\centering\includegraphics[width=\textwidth]{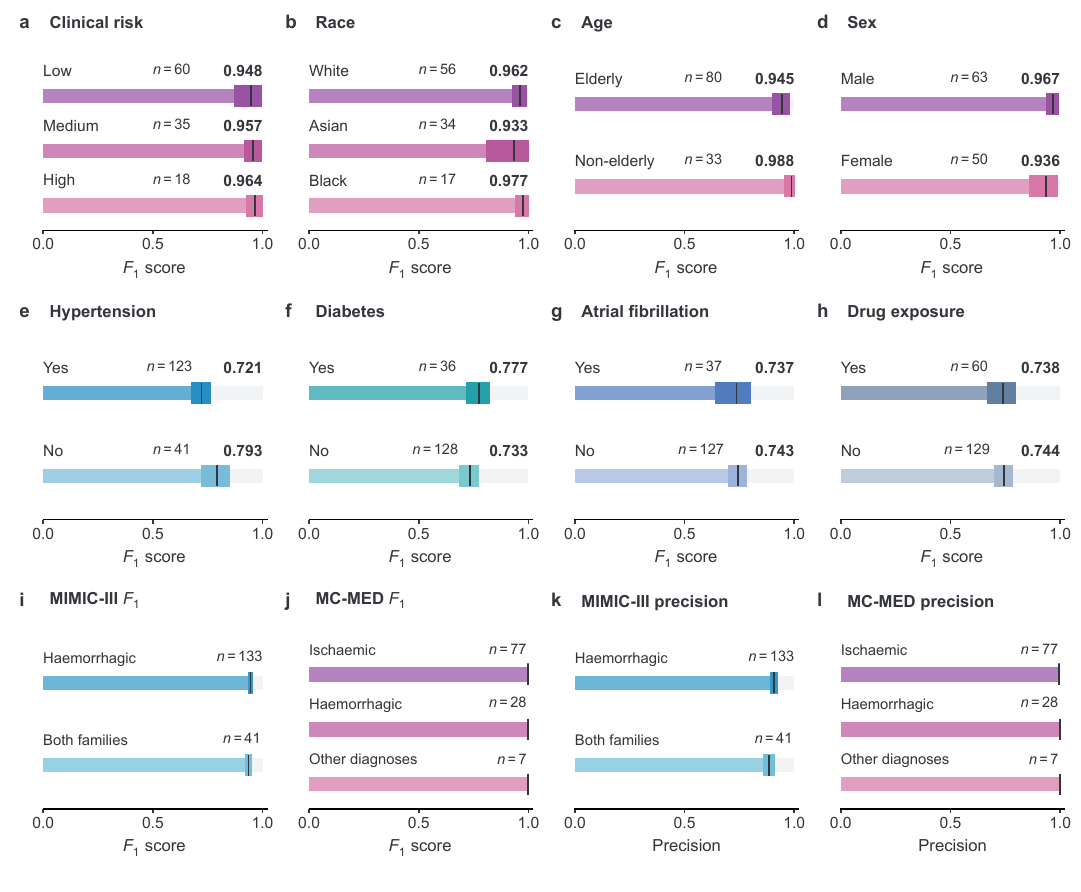}
\caption{Subgroup performance. \textbf{a--d}, $F_1$ of the original MC-MED ensemble by clinical risk, race, age and sex. \textbf{e--h}, $F_1$ of the patient-grouped MIMIC-III ResNet-1D at probability 0.5 by hypertension, diabetes, atrial fibrillation and recent drug exposure. \textbf{i,j}, Diagnostic-code strata of the original 360-min warning analysis, using five-model mean probabilities at cutoff 0.5 in 174 MIMIC-III and 112 MC-MED patients from Table~\ref{tab:cohort-characteristics}. Both coding families and other recorded diagnoses are separate groups. Bar length and vertical marks show point estimates, and darker blocks show 95\% patient-cluster bootstrap intervals (10,000 draws for a--d, 2,000 for e--j). Labels give unique patient counts. \textbf{k,l}, Precision in the same diagnostic groups, with 2,000 patient-cluster bootstrap intervals. Cyan denotes MIMIC-III and magenta MC-MED.}\label{fig:external-subgroup-consistency}
\end{figure}

 The pattern was consistent across clinical-risk, race, age and sex groups in MC-MED, with $F_1$ between 0.93 and 0.99 and overlapping intervals (Fig.~\ref{fig:external-subgroup-consistency}a--d), and across diagnostic-code groups in both cohorts (Fig.~\ref{fig:external-subgroup-consistency}i--l). The diagnostic strata also illustrate a constraint that recurs throughout this design: warning windows made up 88--100\% of each stratum, and two MC-MED strata contained no stable windows at all, so stratum-level $F_1$ chiefly reflects sensitivity.

\subsection*{The signature resides in pulse shape rather than heart rate}

\par
\begin{figure}[H]
\centering\includegraphics[width=\textwidth]{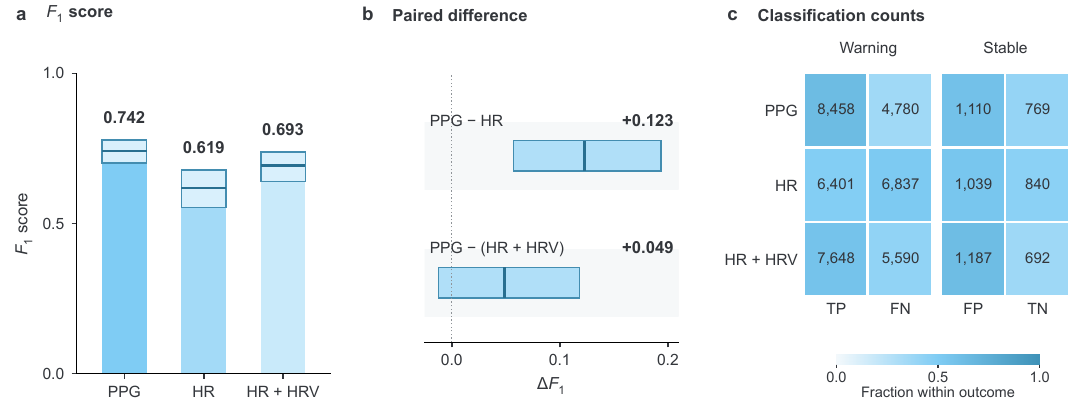}
\caption{Paired morphology, HR and HRV comparison. \textbf{a}, $F_1$. \textbf{b}, Paired $F_1$ differences; blocks show 95\% intervals from 2,000 patient-cluster resamples. \textbf{c}, Confusion counts, coloured by proportion within each true class (13,238 warning and 1,879 stable windows). All models use the same 15,117 windows from 164 MIMIC-III patients and cutoff 0.5. Morphology uses ResNet-1D; HR/HRV use logistic regression with nested tuning.}\label{fig:rate-comparison}
\end{figure}

 We tested whether the information lies in the shape of the pulse or simply in heart rate, which itself changes around many acute neurological and systemic events. On identical windows, morphology exceeded heart rate alone by 0.123 in $F_1$ (95\% CI 0.057--0.194; Fig.~\ref{fig:rate-comparison}). Adding heart-rate variability (HRV) narrowed the difference to 0.049, with an interval that included zero ($-0.012$ to 0.118), indicating that much of the information beyond heart rate overlaps with the autonomic content of beat-to-beat timing. The signal was not confined to patients whose pulse contour would be expected to be distorted by rhythm or drugs: performance was similar with and without recorded atrial fibrillation, recent vasoactive or sedative exposure, hypertension and diabetes (Fig.~\ref{fig:external-subgroup-consistency}e--h).

\subsection*{Pulse haemodynamics after admission stratify the clinical course}

\par If pulse morphology reflects the haemodynamic state surrounding stroke, it should also be informative after the event, when the clinical question becomes which patients will deteriorate. Using only the first two hours of PPG after intensive-care admission or emergency-department arrival, the morphology model distinguished patients whose hospital course worsened or ended in death from those who improved, with $F_1$ of 0.714 in MIMIC-III and 0.691 in MC-MED and recall of about 0.80 in both (Table~\ref{tab:prognosis-performance}). Structured-EHR references, evaluated in their own cohorts over a longer observation window, reached lower values.

\begin{table}[H]
\caption{Performance of PPG and EHR reference models for stroke prognosis prediction.}\label{tab:prognosis-performance}
\centering\npjtablefont\npjtablesetup
\begin{tabularx}{\textwidth}{@{}XYYY@{}}\toprule
Model & Precision & Recall & $F_1$\\\midrule
\multicolumn{4}{@{}l}{\textbf{MIMIC-III}}\\
PPG (ResNet-1D) & \textbf{\ms{0.6486}{0.0290}} & \textbf{\ms{0.8062}{0.1260}} & \textbf{\ms{0.7139}{0.0538}}\\
$\mathrm{CHA}_2\mathrm{DS}_2\text{-}\mathrm{VASc}$ & \ms{0.4908}{0.0327} & \ms{0.6690}{0.0275} & \ms{0.5658}{0.0279}\\
Nwosu--EHR & \ms{0.5138}{0.0452} & \ms{0.5120}{0.0677} & \ms{0.5097}{0.0358}\\
Teoh--EHR & \ms{0.5853}{0.0781} & \ms{0.7361}{0.0821} & \ms{0.6457}{0.0374}\\
Yang--EHR & \ms{0.5566}{0.0728} & \ms{0.6397}{0.0679} & \ms{0.5890}{0.0229}\\
\addlinespace[3pt]
\multicolumn{4}{@{}l}{\textbf{MC-MED}}\\
PPG (ResNet-1D) & \textbf{\ms{0.6292}{0.0948}} & \textbf{\ms{0.8020}{0.1535}} & \textbf{\ms{0.6907}{0.0475}}\\
$\mathrm{CHA}_2\mathrm{DS}_2\text{-}\mathrm{VASc}$ & \ms{0.3811}{0.0000} & \ms{0.5344}{0.0000} & \ms{0.4450}{0.0000}\\
Nwosu--EHR & \ms{0.3845}{0.0176} & \ms{0.4645}{0.0591} & \ms{0.4187}{0.0188}\\
Teoh--EHR & \ms{0.4324}{0.0246} & \ms{0.6226}{0.0920} & \ms{0.5062}{0.0170}\\
Yang--EHR & \ms{0.4186}{0.0334} & \ms{0.5747}{0.1301} & \ms{0.4765}{0.0252}\\
\bottomrule\end{tabularx}\end{table}

\subsection*{Systolic timing before recognition, haemodynamic lability after admission}

\par We next asked which properties of the pulse carry these signals. SHapley Additive exPlanations (SHAP) showed that individual pre-recognition predictions were driven chiefly by crest time ($T_{\mathrm{sp}}$), the interval from pulse onset to systolic peak, expressed both relative to the patient's own baseline and in absolute terms, with smaller and bidirectional contributions from amplitude, timing-ratio and variability descriptors (Fig.~\ref{fig:ppg-morphology-shap}a). Clinical-course predictions drew on a different combination, led by pulse-offset amplitude and timing descriptors (Fig.~\ref{fig:ppg-morphology-shap}b).

\begingroup\setlength{\intextsep}{3pt}
\begin{figure}[H]
    \centering
    \includegraphics[width=0.92\textwidth]
    {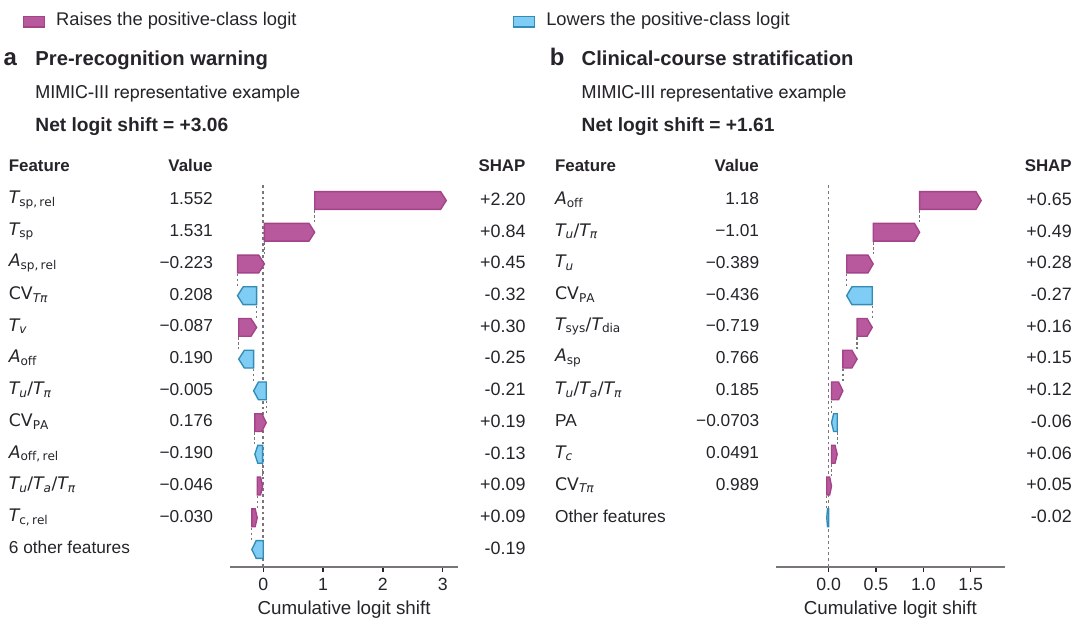}

    \caption{Instance-level SHAP explanations in MIMIC-III. \textbf{a}, Pre-recognition warning. \textbf{b}, Clinical-course stratification. Contributions are to the positive-class logit relative to the model baseline; magenta increases the logit and cyan decreases it.}
\label{fig:ppg-morphology-shap}
\end{figure}
\endgroup

\par
\begin{figure}[H]
    \centering
    \includegraphics[width=\textwidth]{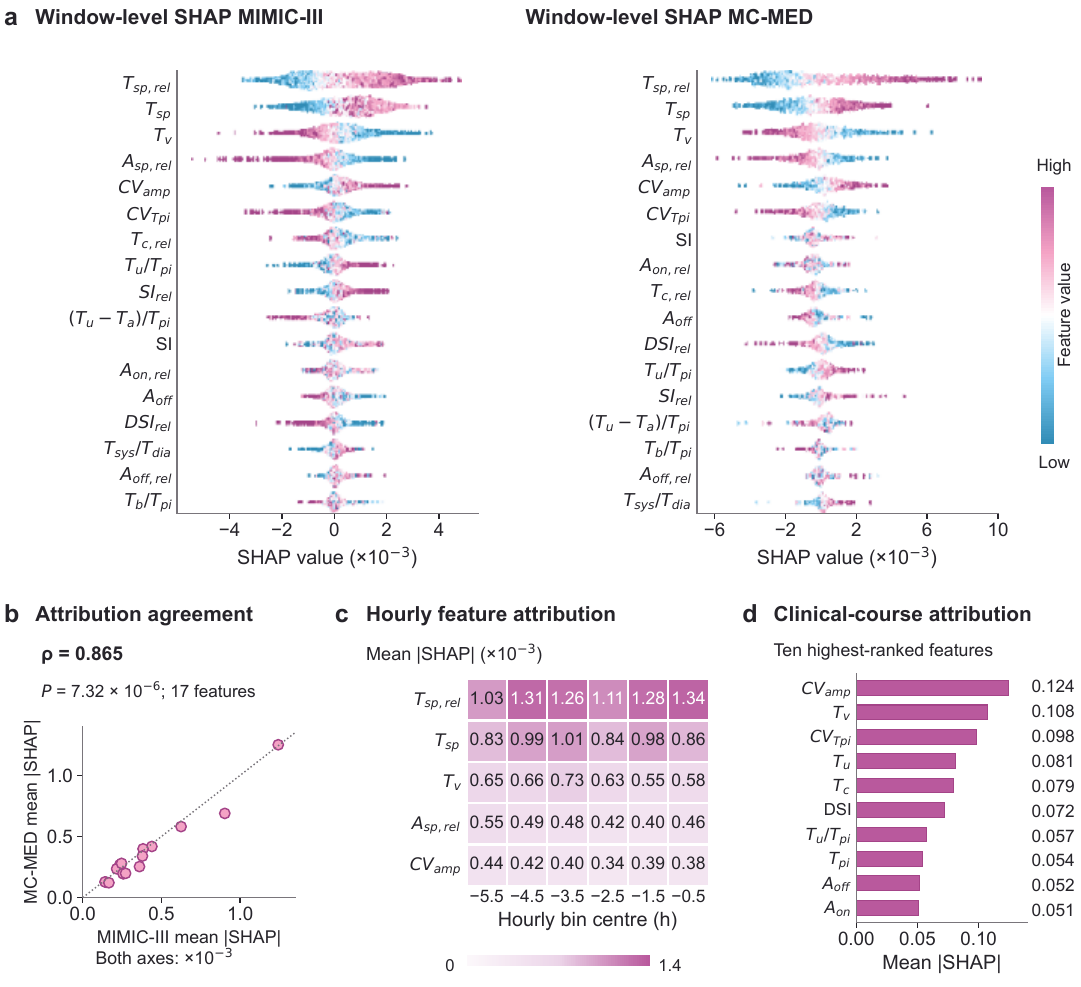}
    \caption{Cohort-level SHAP distributions, cross-cohort agreement and temporal attribution.
\textbf{a}, SHAP beeswarm distributions for the warning model in MIMIC-III and MC-MED. Each plotted point represents an evaluated window; colour indicates the feature value. Features are ordered by cohort-specific mean absolute SHAP value.
\textbf{b}, Agreement of mean absolute attribution across 17 features (Spearman \(\rho=0.865\), two-sided \(P=7.32\times10^{-6}\)). Each point represents one feature; the dashed line indicates equal attribution across cohorts. Both axes are expressed in units of \(10^{-3}\).
\textbf{c}, Mean absolute attribution of the five leading warning features in six hourly bins preceding documented recognition. Axis labels indicate bin centres (\(-5.5\) to \(-0.5\)\,h); cell values and colours share a common \(10^{-3}\) scale. The matrix shows non-monotonic temporal variation in feature attribution.
\textbf{d}, The ten highest-ranked features in the clinical-course model, ordered by mean absolute SHAP value.}
    \label{fig:global-shap}
\end{figure}

 Across all evaluated windows, the same five features carried the greatest attribution in both warning cohorts, led by baseline-relative and absolute crest time and followed by \(T_v\) (onset to the steepest descent), baseline-relative peak amplitude and pulse-amplitude variability (Fig.~\ref{fig:global-shap}a). Attribution ranks were concordant between the intensive-care and emergency-department cohorts (Spearman's \(\rho=0.865\), \(P=7.32\times10^{-6}\)), with the direction of effect shared by 15 of 17 features (Fig.~\ref{fig:global-shap}b). The contribution of crest time was already present six hours before recognition and varied non-monotonically thereafter (Fig.~\ref{fig:global-shap}c). After admission the hierarchy changed, with pulse-amplitude variability, \(T_v\) and pulse-interval variability leading attribution (Fig.~\ref{fig:global-shap}d). The pre-recognition signature is therefore anchored in the timing of systolic ejection and its departure from each patient's own baseline, whereas prognosis after admission is carried by beat-to-beat haemodynamic lability.

\Needspace{5\baselineskip}
\subsection*{Separating crest time from heart rate}

\par
\begin{figure}[H]
\centering\includegraphics[width=\textwidth]{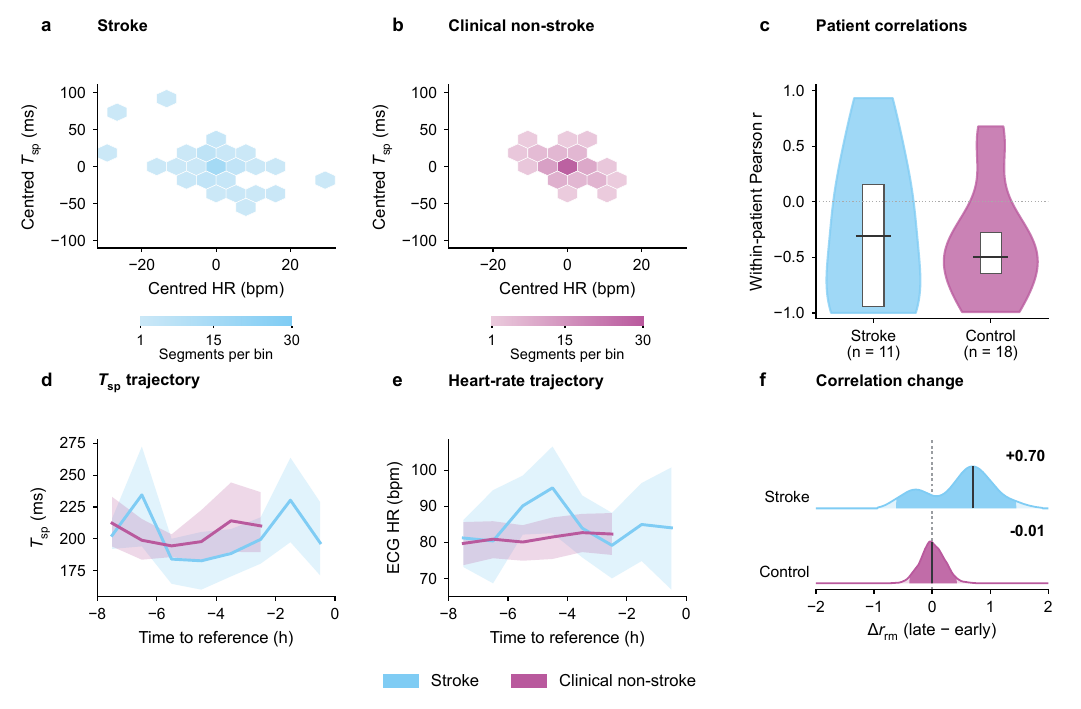}
\caption{$T_{\mathrm{sp}}$--HR association and temporal change in stroke and clinical non-stroke controls. \textbf{a,b}, Within-patient centred HR and crest time; hexagon intensity denotes segment count. \textbf{c}, Patient-level Pearson correlations, with median and interquartile range. \textbf{d,e}, Hourly patient-averaged crest time and ECG HR; shading denotes pointwise 95\% patient-bootstrap intervals. Stroke observations extend to $-0.5$ h and control observations to $-2.5$ h relative to their assigned reference. \textbf{f}, Correlation change $\Delta r_{\mathrm{rm}}$ ($-5$ to $-2$\,h minus $-8$ to $-5$\,h). Densities show 2,000 patient-bootstrap estimates, with the central 95\% shaded darker and observed estimates marked by vertical ticks. Cyan denotes stroke and magenta clinical non-stroke controls. Panels a--e use 60 segments from 20 stroke patients and 101 segments from 21 controls; c uses the 11 stroke and 18 control patients with at least three variable observations. In f, the early and late periods included 5 and 7 stroke patients and 17 and 16 controls, respectively.}\label{fig:crest-hr}
\end{figure}

 Crest time is not a purely vascular quantity, because the systolic upstroke shortens as the cardiac cycle shortens. To separate the two, we analysed synchronous PPG and ECG in 20 MIMIC-III warning patients and 21 propensity-matched controls without stroke (Fig.~\ref{fig:crest-hr}). Crest time shortened with heart rate to the same extent in both groups, by about 12\,ms for each 10-bpm increase, and after adjustment for heart rate it showed no consistent temporal drift in either group (Fig.~\ref{fig:crest-hr}a--e). The chronotropic dependence of crest time is thus shared by patients with and without stroke, and any stroke-related information must lie in departures from it.

\par We then asked whether the coupling itself changes as recognition approaches. In stroke patients, the repeated-measures correlation between crest time and heart rate weakened from $-0.78$ between 8 and 5\,h before recognition to $-0.08$ between 5 and 2\,h before recognition, whereas in controls it remained near $-0.30$ (Fig.~\ref{fig:crest-hr}f). The between-group difference in this change was large but imprecise (0.71; 95\% CI $-0.72$ to 1.54), because only a few stroke patients contributed to each period. A loosening of systolic timing from heart rate is what would be expected if afterload, contractility or vascular tone changed independently of chronotropy; we regard it as a specific hypothesis for prospective testing rather than an established effect.

\subsection*{Specificity in high-risk patients without stroke}

\par A haemodynamic signature of stroke must be distinguishable from the fluctuations of other acutely ill patients. We applied the warning models to independent high-risk controls, aged 65 years or older with hypertension or diabetes and monitored for at least 24\,h without documented stroke, and examined both individual windows and persistent alerts, defined as at least five consecutive positive windows (Methods).

\par
\begin{figure}[H]
    \centering
    \includegraphics[width=\textwidth]{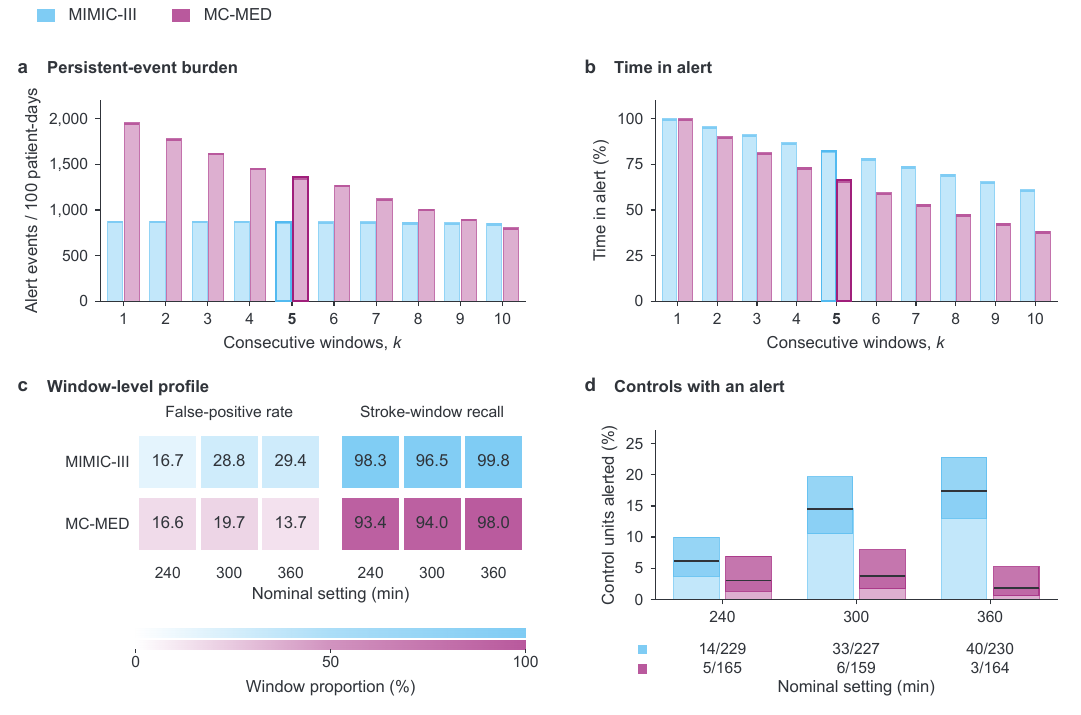}
    \caption{Persistent alerts in non-stroke controls. \textbf{a,b}, Alert events per 100 patient-days and time in alert for one to ten consecutive windows; outlined columns mark the five-window rule. \textbf{c}, Window false-positive rate and stroke-window recall (TPR). Cell values are percentages; both cohort colour scales span 0--100\%. \textbf{d}, Proportion of controls with a five-window alert; darker blocks show 95\% Wilson intervals and black horizontal marks show estimates. Counts below the axis are alerted/total control units. Cyan denotes MIMIC-III and magenta MC-MED.}
\label{fig:alert-burden}
\end{figure}

 Single positive windows were common in controls: across horizons, 17--29\% of control windows in MIMIC-III and 14--20\% in MC-MED were classified as positive (Table~\ref{tab:false-alert-burden}). Requiring persistence changed the picture. Under the five-window rule, alerts occurred in 6--17\% of MIMIC-III controls and 2--4\% of MC-MED controls, while sensitivity for warning windows in stroke patients remained above 0.93; lengthening the requirement further reduced both alert events and time in alert, most steeply in MC-MED (Fig.~\ref{fig:alert-burden}). Part of the lower MC-MED rate reflects shorter monitoring, with fewer evaluable windows per control (Methods). Transient positive windows are thus frequent in patients without stroke, whereas sustained positivity is considerably less common, making temporal persistence the natural axis on which a pre-stroke signature can be separated from background haemodynamic variability.

\begin{table}[H]
\caption{False-alert burden among high-risk non-stroke controls across nominal temporal settings.}
\label{tab:false-alert-burden}
\centering
\npjtablefont
\setlength{\tabcolsep}{2.0pt}
\renewcommand{\arraystretch}{1.10}

\begin{tabularx}{\textwidth}{@{}
    C{1.15cm}
    C{1.45cm}
    C{1.70cm}
    C{1.60cm}
    Y
    Y
    Y
@{}}
\toprule
\npjtablehead{Setting} &
\shortstack[c]{\npjtablehead{Control}\\\npjtablehead{units, \(n\)}} &
\shortstack[c]{\npjtablehead{Retained}\\\npjtablehead{windows, \(n\)}} &
\shortstack[c]{\npjtablehead{Persistent}\\\npjtablehead{alerts, \(n\)}} &
\shortstack[c]{\npjtablehead{Window}\\\npjtablehead{FPR}} &
\shortstack[c]{\npjtablehead{Persistent-}\\\npjtablehead{alert rate}} &
\shortstack[c]{\npjtablehead{Stroke-window}\\\npjtablehead{TPR}} \\
\midrule

\multicolumn{7}{@{}l}{\textbf{a, MIMIC-III}} \\[-1pt]
240\,min & 229 & 2,997 & 14 & 0.1672 & 0.0611 & 0.9833 \\
300\,min & 227 & 2,954 & 33 & 0.2881 & 0.1454 & 0.9647 \\
360\,min & 230 & 2,910 & 40 & 0.2938 & 0.1739 & 0.9981 \\

\addlinespace[3pt]
\midrule
\addlinespace[1pt]

\multicolumn{7}{@{}l}{\textbf{b, MC-MED}} \\[-1pt]
240\,min & 165 & 1,257 & 5 & 0.1663 & 0.0303 & 0.9341 \\
300\,min & 159 & 1,275 & 6 & 0.1969 & 0.0377 & 0.9401 \\
360\,min & 164 & 1,284 & 3 & 0.1371 & 0.0183 & 0.9804 \\

\bottomrule
\end{tabularx}

\vspace{3pt}

\end{table}

\subsection*{Patient-level alerts and monitoring coverage}

\par
\begin{figure}[H]
    \centering
    \includegraphics[width=\textwidth]{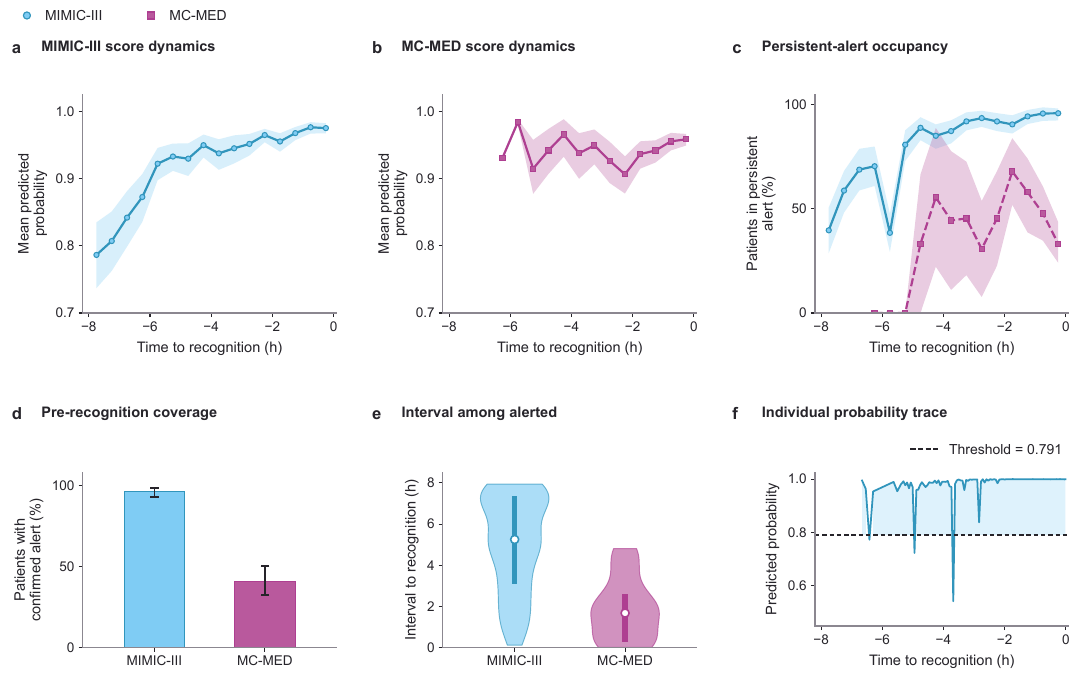}
    \caption{Available-waveform alert coverage and conditional intervals to documented recognition.
\textbf{a,b}, Patient-equal mean probabilities in MIMIC-III and MC-MED, with pointwise 95\% patient-cluster bootstrap intervals. Patient support varies by bin; bins containing one patient are descriptive.
\textbf{c}, Persistent-alert occupancy among patients contributing to each time bin, with bootstrap intervals.
\textbf{d}, Patients with a confirmed alert: 168/174 (96.6\%) and 46/112 (41.1\%); whiskers are Wilson 95\% intervals. All patients remain in the denominator, including 62 MC-MED patients with fewer than five windows.
\textbf{e}, Intervals among alerted patients, expressed in hours. White points and thick bars denote medians and interquartile ranges: 5.25 [3.24--7.22] and 1.68 [0.42--2.49].
\textbf{f}, One MIMIC-III trace; the dashed line marks the frozen threshold 0.791 and shading identifies above-threshold output. Confirmation occurs at the end of the fifth consecutive window with probability greater than 0.791. For patients with multiple recordings or anchors, the earliest confirmed relative time is retained. Cyan circles denote MIMIC-III and magenta squares MC-MED. }

    \label{fig:lead-time}
\end{figure}

 Applied to all available waveform segments, the five-window rule raised an alert before recognition in 168 of 174 MIMIC-III patients (96.6\%) but in only 46 of 112 MC-MED patients (41.1\%) (Fig.~\ref{fig:lead-time}). Among alerted patients, the first confirmed alert preceded documented recognition by a median of 5.25\,h in MIMIC-III and 1.68\,h in MC-MED. These intervals are conditional on an alert being raised and are measured to documentation rather than to onset; they should not be read as prospective lead times.

\par
\begin{figure}[H]
\centering\includegraphics[width=\textwidth]{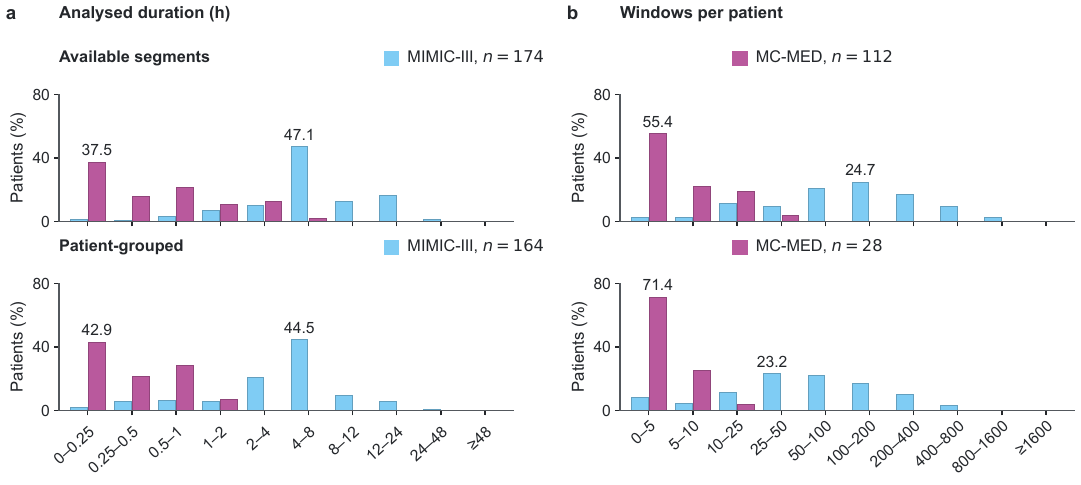}
\caption{Patient-level waveform coverage. \textbf{a}, Analysed duration. \textbf{b}, Windows per patient. The upper row shows the available-segment alert population; the lower row shows the patient-grouped population. Cyan and magenta bars show MIMIC-III and MC-MED, respectively, on a common 0--80\% scale. Labels identify the largest bin in each cohort distribution. Duration merges overlapping intervals within patients. Unequal-width bins are displayed as categories.}
\label{fig:monitoring}
\end{figure}

 The difference in coverage between the cohorts reflected the monitoring available rather than the quality of the signal. Most MC-MED patients had less than an hour of analysable PPG before recognition (median 0.44\,h, compared with 7.39\,h in MIMIC-III), and 62 of the 112 had fewer than five windows, too few for the persistence rule to operate (Fig.~\ref{fig:monitoring}). Signal quality, by contrast, was high in both cohorts, with median patient-level signal-quality indices of 0.87 and 0.93, and nearly all attempted recordings were processed successfully (Table~\ref{tab:signal-quality-retention}).

\begin{table}[H]
\captionsetup{singlelinecheck=false,justification=raggedright}
\caption{PPG signal quality, waveform retention and processing success across datasets.}
\label{tab:signal-quality-retention}
\centering
\npjtablefont
\setlength{\tabcolsep}{5pt}
\renewcommand{\arraystretch}{1.20}

\begin{tabularx}{\textwidth}{@{}
    L{8.50cm}
    Y
    Y
@{}}
\toprule
\npjtablehead{Measure} &
\npjtablehead{MIMIC-III} &
\npjtablehead{MC-MED} \\
\midrule

\multicolumn{3}{@{}l}{\textbf{a, Signal quality}} \\[-1pt]
Evaluated patients, \(n\)
& 312
& 158 \\
SQI, mean (s.d.)
& 0.85 (0.11)
& 0.90 (0.08) \\
SQI, median [IQR]
& 0.87 [0.79--0.93]
& 0.93 [0.86--0.96] \\

\addlinespace[5pt]
\multicolumn{3}{@{}l}{\textbf{b, Waveform availability and processing}} \\[-1pt]
Warning hourly retention
& 98.4--98.8\%
& 82.9--98.8\% \\
Date-associated time (DAT) retention
& 67.61\%
& 44.59\% \\
Clinical-course duration retention
& 82.41\%
& 91.24\% \\
Processing success
& 98.73\%
& 100.00\% \\

\bottomrule
\end{tabularx}

\vspace{3pt}

\end{table}

\FloatBarrier
\section*{Discussion}

\par Stroke is usually studied as the endpoint of slowly accumulating vascular risk. Our findings bring a different timescale into view. In two independent care settings, the morphology of the routinely recorded pulse differed systematically in the hours before stroke recognition from the same patients' earlier state. The difference was carried mainly by the timing of systolic ejection, was expressed relative to each patient's own baseline, ranked concordantly across cohorts and transferred without retraining from intensive care to the emergency department. The same waveform, recorded early after admission, stratified the subsequent clinical course, with attribution shifting from systolic timing to beat-to-beat haemodynamic lability. Pulse morphology therefore carries a haemodynamic imprint of the period surrounding stroke, and the pulse oximeter, present at every monitored bedside and on hundreds of millions of wrists, is in principle an instrument for observing it.

\par The signature is anchored in a feature whose physiology is well characterised. Crest time, the interval from pulse onset to systolic peak, is shaped by the rate and duration of left-ventricular ejection, the compliance of the aorta and conduit arteries and the arrival of waves reflected from the periphery; arterial stiffening, changes in reflection, inotropy and vasoactive drugs all alter it \cite{Allen2007PPG,Charlton2022Hemodynamics,Takazawa1998APG}. It also shortens as heart rate rises, and our synchronous ECG analysis shows that this dependence is the same in patients with and without stroke. A stroke-related signal in crest time must therefore lie in departures from the expected chronotropic relationship, and the loosening of the crest-time--heart-rate coupling before recognition in stroke patients, absent in controls, is the kind of departure that neurocardiac physiology predicts. Pulse timing should dissociate from heart rate whenever afterload, contractility or vascular tone change independently of chronotropy: in the sympathetic surge that follows insular or brainstem injury; in the Cushing physiology of rising intracranial pressure, in which hypertension accompanies relative bradycardia; and in neurogenic myocardial stunning, in which contractility falls while rate rises \cite{Qureshi2008AHR,Scheitz2022StrokeHeart}. 

\par Several features of the data suggest that the signature is personal. Baseline-relative crest time was the most influential feature in both cohorts, yet pooled patient-grouped discrimination was low. These observations fit together if the change that precedes stroke is small relative to differences in absolute pulse morphology between individuals, which reflect age, arterial stiffness, peripheral perfusion, probe site and medication. A signature expressed as change within each patient is diluted when windows from many patients are placed on a common scale, much as the electrocardiographic diagnosis of myocardial ischaemia rests on change from a patient's own baseline more than on absolute values. Clinicians read monitors the same way: a heart rate of 95\,bpm or a slightly damped pulse matters because it differs from the same patient's pattern an hour earlier. We did not measure within-patient discrimination directly, and a weak signal would produce the same pooled result, so this interpretation must be tested explicitly. If it holds, it has a clear implication for biomarker design: models should be built around each individual's trajectory rather than around population thresholds.

\par After admission, the dominant features changed from systolic timing to beat-to-beat amplitude variability, the timing of the steepest descent and pulse-interval variability, all of which index haemodynamic lability. This shift has a physiological rationale. Cerebral autoregulation is impaired after stroke \cite{Intharakham2019Autoregulation}, so fluctuations in systemic pressure are transmitted more directly to injured tissue, and systolic blood-pressure variability after intracerebral haemorrhage independently predicts death and major disability \cite{Manning2014BPV}. After aneurysmal subarachnoid haemorrhage, delayed cerebral ischaemia develops over days in patients whose blood pressure is actively manipulated \cite{Macdonald2014DCI}. That two hours of pulse morphology stratified the subsequent course suggests that the lability visible in the peripheral pulse reflects, at least in part, the haemodynamic vulnerability of the injured brain.

\par The cohorts also sample distinct haemodynamic contexts. Every MIMIC-III warning patient carried an admission-level haemorrhagic code, and many admissions contained more than one recognition anchor, the profile of a neurocritical care population in which new or worsening deficits arise in patients already admitted with intracerebral or subarachnoid haemorrhage. In such patients blood pressure is actively lowered or deliberately raised, intracranial pressure may be rising, and vasoactive, sedative and osmotic therapies reshape the pulse. MC-MED captured older emergency patients with predominantly ischaemic diagnoses, in whom the pre-recognition interval more often represents diagnostic delay in a stroke that has already begun; about 9\% of strokes are initially missed in the emergency department, more often when symptoms are mild, transient or non-specific \cite{Tarnutzer2017EDMisdiagnosis}. 

\par Moving from signature to surveillance will depend on how the signal behaves over time within individual patients. Temporal persistence was the most effective filter, suppressing most alerts arising from transient fluctuations in controls while preserving sensitivity in stroke patients. Coverage, in turn, was limited by the monitoring available rather than by signal quality, as the brief recordings of the emergency-department cohort showed. The most plausible first application is therefore a prompt for focused neurological examination in populations with high pretest probability, such as patients after cardiac surgery or transcatheter procedures, after intracerebral or subarachnoid haemorrhage, or with atrial fibrillation and interrupted anticoagulation, evaluated against a prespecified alert budget that respects the existing burden of monitor alarms \cite{Drew2014AlarmFatigue}. The larger opportunity lies outside hospital, where most strokes occur and where wearable and contactless sensors already record the same waveform \cite{Charlton2023WearablePPGRoadmap,Huang2023VCPM}; long personal baselines from such devices are precisely what a within-patient signature requires.

\par Further limitations apply. Both datasets are retrospective and come from single health systems in the United States, and MIMIC-III predates current stroke practice. Pulse oximetry is affected by skin pigmentation \cite{Sjoding2020PulseOx}; our race-stratified estimates were similar but came from small groups with $F_1$ values near ceiling. Vasoactive and sedative exposure was captured only as a binary six-hour indicator, and fluid balance, temperature, peripheral perfusion, probe site and motion were not modelled, although each alters pulse morphology. Labels were derived from clinical notes by an LLM-assisted procedure with physician adjudication; agreement with adjudicated times was close in an independently reviewed sample, but documentation practices differ between institutions and eras.

\section*{Methods}

\subsection*{Study design, data sources and ethics}

\par We conducted a retrospective cross-dataset study of PPG morphology for two stroke-related tasks: warning-state classification relative to a physician-adjudicated temporal anchor for documented in-hospital stroke recognition, and hospitalisation-level clinical-course stratification after a stroke-associated admission. We developed and evaluated models internally in MIMIC-III, separating the historical and strict analyses, then applied the frozen models to MC-MED. Model parameters, preprocessing transformations and decision thresholds determined during MIMIC-III development were applied unchanged to MC-MED. Reporting was guided by TRIPOD+AI and complementary guidance for clinical artificial-intelligence studies \cite{Collins2024TRIPODAI,Norgeot2020MICLAIM}.

\par We used two de-identified waveform-linked clinical resources: the MIMIC-III Clinical Database with its Waveform Database Matched Subset v1.0 \cite{Johnson2016MIMICIII,Moody2020MIMICIIIWaveform} and MC-MED v1.0.0 \cite{Kansal2025MCMED,Kansal2025MCMEDDataset}. MIMIC-III contains critical-care data from Beth Israel Deaconess Medical Center and links continuous physiological recordings to corresponding clinical records. MC-MED contains multimodal data from monitored adult emergency-department encounters at Stanford Health Care, including structured EHR data and continuous physiological waveforms.

\par MIMIC-III was released following institutional review by Beth Israel Deaconess Medical Center and the Massachusetts Institute of Technology, with the requirement for individual informed consent waived for this de-identified research resource \cite{Johnson2016MIMICIII}. MC-MED was approved by the Stanford University Institutional Review Board under protocol no.~58581 and released for retrospective research subject to the applicable data-use requirements \cite{Kansal2025MCMED,Kansal2025MCMEDDataset}. Access to both resources complied with the applicable credentialing, training and data-use requirements. The present study involved no prospective recruitment or intervention.

\subsection*{Cohort definition and clinical outcomes}

\par We linked waveform records to the corresponding patient or hospitalisation timeline. Warning cohorts included patients with an in-hospital stroke and usable PPG before the physician-adjudicated documented-recognition anchor. Temporal linkage, waveform availability and signal-quality processing yielded 176 MIMIC-III and 154 MC-MED patients. The subsequent patient-grouped evaluation assigned all records from a patient to one fold.

\par High-risk non-stroke controls were aged $\geq65$ years, had hypertension or diabetes, received monitored care for at least 24\,h and met waveform-availability, quality and retention criteria. We applied the stroke-trained models to their monitored windows to quantify false-positive windows and persistent-alert burden.

\par Clinical-course cohorts included stroke-associated hospitalisations with adjudicated final status. The index-care anchor was ICU admission in MIMIC-III and emergency-department arrival in MC-MED. The original sequence-model input comprised the first 120\,min of usable, aligned PPG. Waveform-linked source records numbered 1,810 and 3,472, respectively. Structured-EHR references used separate visit-level cohorts of 1,932 MIMIC-III visits from 1,911 patients and 1,182 MC-MED visits from 1,122 MRN groups, with inputs defined below. Outcome filtering, waveform eligibility and the observation interval define each model-specific subset.

\par Diagnostic groups in Table~\ref{tab:cohort-characteristics} use the same 176 and 154 warning patients and 1,810 and 3,472 clinical-course source records as the other characteristics. MIMIC-III groups use the index-admission ICD-9 families 433/434/436 (ischaemic/occlusive) and 430/431/432 (haemorrhagic); MC-MED groups use recorded I63 and I60--I62 diagnoses, respectively, and overlapping families are reported separately. Other MC-MED diagnoses comprised six patients with subdural haematoma and one with carotid stenosis in the warning cohort, and 810 traumatic intracranial-injury, 286 transient-ischaemic-attack and 23 carotid-stenosis records in the clinical-course cohort. Retained and excluded warning candidates (176 retained and 158 excluded in MIMIC-III; 154 retained and 171 excluded MRN groups in MC-MED) were compared with Welch $t$ tests or Fisher exact tests without multiplicity adjustment.

\subsection*{Clinical label ascertainment}

\par The temporal reference for the warning analysis was a final physician-adjudicated timestamp derived from the available documentation of an in-hospital stroke. An LLM-assisted information-extraction procedure identified candidate times in de-identified clinical notes, following our preceding in-hospital stroke study \cite{Liu2026InHospitalStroke,Yang2022EHRLLM,Agrawal2022ClinicalLLM}. The extraction procedure used Gemini 3 Pro Preview (Google) with a predefined prompt. Warning-label extraction was performed in November 2025 and clinical-course extraction in April 2026. Inputs to the LLM were limited to a de-identified record key, note timestamp and note text; PPG waveforms, PPG-derived features, structured prediction variables and prediction outputs were not provided.

\par The extraction procedure identified acute stroke or cerebrovascular events and newly documented neurological deficits. It prioritised explicit symptom-onset times, using discovery time, last-known-well information and note metadata as supporting evidence. For acute events described without an explicit onset time, note timestamps provided candidate times for clinical adjudication. The final adjudicated time, termed the documented-recognition anchor, aligned the waveform observations to the clinical event record.

\par We checked candidate timestamps against note and admission timing and removed duplicate waveform--timestamp mappings. Two neurologists independently reviewed the temporal evidence. The adjudicated anchor, $T_i$ for patient $i$, defined window construction and evaluation. In a stratified sample of 100 independently reviewed cases, 95.0\% of LLM-assisted candidate timestamps were within $\pm15$\,min of the physician-adjudicated time.

\par For clinical-course stratification, the LLM-assisted procedure generated candidate hospitalisation statuses from \{improved, stable, worsened, deceased\} and returned an unavailable status when documentation was insufficient. Candidate statuses were checked against the clinical record and then clinically adjudicated. The binary outcome was defined as
\[
y_i^{\mathrm{course}}
=
\mathbb{I}\!\left(
r_i \in \{\mathrm{worsened},\mathrm{deceased}\}
\right),
\]

where \(r_i\) denotes the final adjudicated hospitalisation status. Records adjudicated as improved were assigned \(y_i^{\mathrm{course}}=0\), records adjudicated as worsened or deceased were assigned \(y_i^{\mathrm{course}}=1\), and records with stable or unavailable final status were excluded. The final clinical-course label could incorporate information recorded after the respective predictor-observation interval and was not included among model inputs.

\subsection*{Warning-window definition and PPG morphology representation}

\par For patient \(i\), let \(T_i\) denote the physician-adjudicated documented-recognition anchor and \(\tau<T_i\) a candidate evaluation time. Predictors were restricted to information available at or before \(\tau\). Anchor-relative time and the nominal interval to the anchor were defined as

\[
t_{i,\tau}=\tau-T_i,
\qquad
L_{i,\tau}=T_i-\tau.
\]

The nominal temporal settings were \(T_w\in\{240,300,360\}\) min. A stable look-back limit of 480\,min was used together with a 15-min transition buffer around each stable--warning boundary (\(\Delta_{\mathrm{pre}}=15\) min) and a separate 15-min blind zone adjacent to the anchor (\(\Delta_0=15\) min). The warning label was

\[
y_{i,\tau}^{(T_w)}=
\begin{cases}
1,
&
-\left(T_w-\Delta_{\mathrm{pre}}\right)
\leq t_{i,\tau}< -\Delta_0,\\[4pt]
0,
&
-480\leq t_{i,\tau}
\leq -\left(T_w+\Delta_{\mathrm{pre}}\right),\\[4pt]
\text{excluded},
&
\text{otherwise}.
\end{cases}
\]

\par Observations within \(\Delta_{\mathrm{pre}}\) of the nominal stable--warning boundary and observations from 15\,min before the adjudicated anchor onwards were excluded under the prespecified temporal protocol. The transition buffer reduced ambiguity at the class boundary, whereas the blind zone prevented the warning model from using physiology immediately adjacent to the adjudicated reference.

\par We aligned each PPG source file to its clinical timeline. Warning samples were mapped to absolute timestamps and then expressed relative to $T_i$. Clinical-course sequence inputs comprised the first 120\,min. Preprocessing and window construction operated separately within each source recording.

\par Beat-level morphology was extracted using pyPPG \cite{Goda2024pyPPG}. Prior feature selection removed variables with Cohen's $d\leq0.05$ and redundant variables from pairs with $|r|>0.80$, yielding a fixed set of 17 timing, amplitude, contour and variability descriptors \cite{Liu2026InHospitalStroke}. The patient-grouped analysis used this fixed representation and fitted imputation and scaling within each MIMIC-III training fold. These transformations were applied unchanged to MC-MED. SQI and retention summaries characterised recording availability separately from the prediction inputs.

\par In the patient-grouped analysis, baseline-relative descriptors used the median $m_{i,\mathrm{base}}$ of the most recent 500 beats from the same source in the 30--15\,min interval before the input-window start:
\[
F_{i,\mathrm{rel}}(\tau)=\frac{x_{i,\tau}-m_{i,\mathrm{base}}}{|m_{i,\mathrm{base}}|}.
\]
A non-finite median or $|m_{i,\mathrm{base}}|\leq10^{-6}$ produced a missing value and its indicator. Windows required all 500 baseline beats in chronological sequence, with no interbeat gap exceeding 5\,s. The original-window feature files instead used the median of the first $\max\{5,\lfloor0.1N\rfloor\}$ beats in each time-sorted record of $N$ beats and divided the displacement by that signed median; a zero or missing median was replaced by $10^{-6}$. The local field $T_{\mathrm{sys}}=(sp-on)/f_s$ measures beat-relative crest time.

\par
\begin{figure}[!htbp]
\centering
\includegraphics[width=\textwidth]{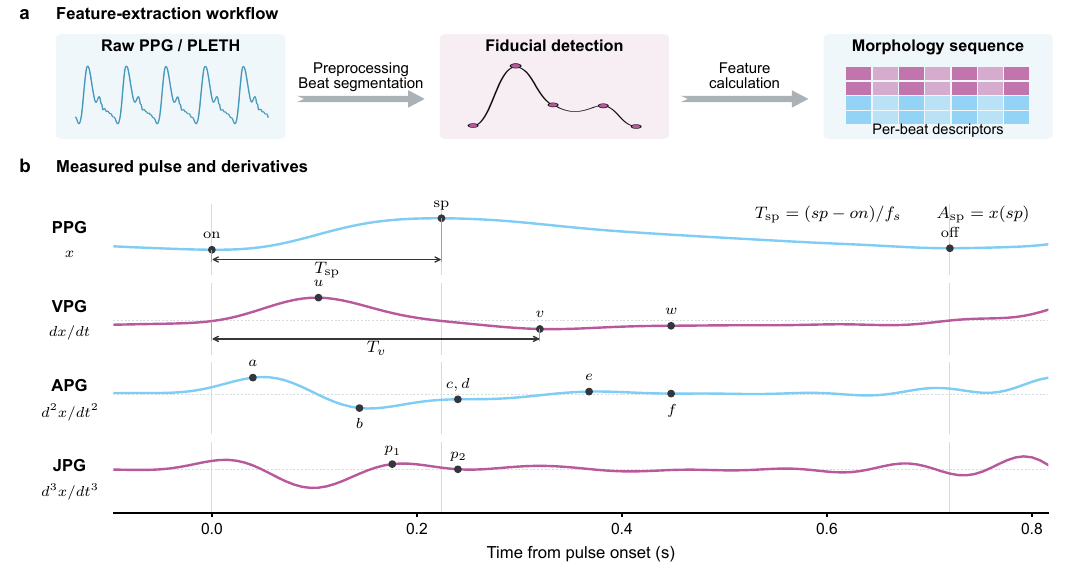}
\caption{PPG feature extraction. \textbf{a}, Schematic processing workflow based on pyPPG~\cite{Goda2024pyPPG}. \textbf{b}, A measured pulse, velocity (VPG), acceleration (APG) and third derivative (JPG) on a common time axis. Landmarks use their detected sample positions; coincident c,d points share a label. Onset, peak and offset are $on$, $sp$ and $off$; arrows span the onset-to-peak interval $T_{\mathrm{sp}}$ and onset-to-VPG-valley interval $T_v$. $f_s$ is sampling frequency and $x$ the filtered PPG. Table~\ref{tab:inputs} defines all 17 descriptors.}
\label{fig:ppg-methods}
\end{figure}

 Table~\ref{tab:inputs} groups the 17 morphology descriptors, and Fig.~\ref{fig:ppg-methods} illustrates the extraction steps using a measured PPG segment.

\begin{table}[!htbp]
\caption{Fixed 17-channel PPG morphology input.}\label{tab:inputs}\label{tab:ppg_feature_definitions}
\centering\small
\setlength{\tabcolsep}{5pt}\renewcommand{\arraystretch}{1.10}
\begin{tabularx}{\textwidth}{@{}>{\raggedright\arraybackslash}p{0.20\textwidth}>{\raggedright\arraybackslash}X@{}}
\toprule
\textbf{Feature} & \textbf{Definition and physiological description} \\
\midrule
\addlinespace[3pt]
\multicolumn{2}{@{}l}{\textbf{Time-domain morphology (4 channels)}} \\
\addlinespace[1pt]
$T_{\mathrm{sp}}$ & Onset-to-peak interval; duration of the systolic upstroke. \\
$\mathrm{SI}$ & Pulse amplitude divided by crest time; mean systolic upstroke slope. \\
$A_{\mathrm{off}}$ & PPG amplitude at pulse offset; end-of-pulse signal level. \\
$T_{\mathrm{sys}}/T_{\mathrm{dia}}$ & Onset-to-peak duration divided by peak-to-offset duration; relative lengths of the rising and falling phases. \\
\addlinespace[3pt]
\multicolumn{2}{@{}l}{\textbf{Derivative-based timing (4 channels)}} \\
\addlinespace[1pt]
$T_u/T_{\pi}$ & Pulse-normalised onset-to-VPG maximum time; timing of maximum upstroke velocity. \\
$T_b/T_{\pi}$ & Pulse-normalised onset-to-APG b time; timing of early upstroke deceleration. \\
$T_v$ & Onset-to-VPG minimum time; timing of the maximum descending slope. \\
$(T_u-T_a)/T_{\pi}$ & Pulse-normalised delay between APG a and VPG u; relative timing of maximum upstroke acceleration and velocity. \\
\addlinespace[3pt]
\multicolumn{2}{@{}l}{\textbf{Short-term variability (2 channels)}} \\
\addlinespace[1pt]
$\mathrm{CV}_{T_{\pi}}$ & Five-beat pulse-interval coefficient of variation; short-term rhythm variability. \\
$\mathrm{CV}_{\mathrm{PA}}$ & Five-beat pulse-amplitude coefficient of variation; beat-to-beat amplitude variability. \\
\addlinespace[3pt]
\multicolumn{2}{@{}l}{\textbf{Within-patient relative deviation (7 channels)}} \\
\addlinespace[1pt]
$T_{\mathrm{sp,Rel}}$ & Baseline-relative crest time; within-patient change in systolic upstroke duration. \\
$A_{\mathrm{sp,Rel}}$ & Baseline-relative systolic-peak amplitude; within-patient change in peak signal level. \\
$\mathrm{SI}_{\mathrm{Rel}}$ & Baseline-relative upstroke slope; within-patient change in pulse-rise geometry. \\
$\mathrm{DSI}_{\mathrm{Rel}}$ & Baseline-relative composite timing ratio, $(T_f+T_{p2})/T_{\pi}$; within-patient change in derivative-landmark timing. \\
$T_{c,\mathrm{Rel}}$ & Baseline-relative onset-to-APG c time; within-patient change in the timing of this acceleration landmark. \\
$A_{\mathrm{off,Rel}}$ & Baseline-relative pulse-offset amplitude; within-patient change in end-of-pulse signal level. \\
$A_{\mathrm{on,Rel}}$ & Baseline-relative pulse-onset amplitude; within-patient change in onset signal level. \\
\bottomrule
\end{tabularx}
\end{table}

\subsection*{PPG model development}

\subsubsection*{Warning-state classification model}

\par The warning model used a one-dimensional residual neural network (ResNet-1D) to process ordered PPG morphology sequences \cite{He2016ResNet}. The original development code grouped local recording identifiers. The patient-grouped analysis grouped by MIMIC-III SUBJECT\_ID across all linked admissions, using five outer folds and training-partition model selection. Historical training used weighted cross-entropy with a positive-class weight of 3.0 and Adam optimisation with a learning rate of \(10^{-3}\) and weight decay of \(10^{-4}\).

\par In the historical branch, the training epoch was selected by validation AUC, with validation \(F_2\) used to resolve ties. At each validation epoch, we evaluated decision thresholds from 0.50 to 1.00 in increments of 0.01. Thresholds were ranked by \(F_2\), followed sequentially by \(F_1\), recall, AUC and threshold value as tie-breakers. This procedure prioritised sensitivity during threshold selection while retaining \(F_1\) as the primary reported threshold-dependent performance metric. The five cross-validation models, preprocessing transformations and validation-derived thresholds were then evaluated in MC-MED without further fitting.

\par The patient-grouped ResNet used 500 beats with 17 values and 17 missing indicators per beat, giving 34 input channels and eight residual blocks. Five outer folds and three inner folds selected learning rate from $\{3\times10^{-4},10^{-3}\}$ and weight decay from $\{10^{-4},10^{-3}\}$, with batch size 128, at most 30 epochs and patience five. Positive-class weights equalled training negatives divided by positives (0.1252--0.1516 across outer refits). Five seeds were run, with imputation and scaling fitted in the corresponding training partition. Calibration used inner out-of-fold scores within each outer-training fold.

\subsubsection*{Clinical-course PPG model}

\par Clinical-course stratification used the same ResNet-1D model family as a separate hospitalisation-level prediction task. Each eligible hospitalisation contributed an ordered beat-level PPG morphology sequence restricted to the original 120-min predictor-observation interval. Training sampled one random 2,048-beat subsequence per eligible sequence. Evaluation scanned all available beats within that 120-min interval using 2,048-beat windows and a stride of 1,024 beats.

\par MIMIC-III development used patient-grouped five-fold cross-validation, with all hospitalisations from the same patient assigned to one fold. Window-level probabilities were summarised using candidate maximum, mean and consecutive-window aggregation rules; the aggregation rule and corresponding decision threshold were selected from MIMIC-III validation data. Training used Adam optimisation with a learning rate of \(10^{-4}\), weight decay of \(10^{-4}\), batch size 16, a maximum of 500 epochs and early stopping after 20 consecutive epochs without validation improvement. Class imbalance was addressed by inverse-frequency weighting. For \(C\) classes and \(n_c\) training observations in class \(c\),

\[
w_c=
\frac{\sum_{j=1}^{C}n_j}{C n_c},
\]

and the weighted cross-entropy objective was

\[
\mathcal{L}_{\mathrm{CE}}
=
-\sum_{c=1}^{C}w_c y_c\log p_c,
\]

where \(y_c\) is the one-hot class indicator and \(p_c\) the predicted class probability. The cross-validation models, preprocessing transformations, aggregation rule and thresholds selected in MIMIC-III were evaluated in MC-MED without further fitting.

\subsection*{Structured-EHR reference models}

\par Four structured-EHR approaches were included as contextual clinical references: an available-component \(\mathrm{CHA}_2\mathrm{DS}_2\text{-}\mathrm{VASc}\) score \cite{Lip2010CHA2DS2VASc,Kim2017CHA2DS2VASc} and machine-learning models based on the information domains described by Nwosu et al. \cite{Nwosu2019StrokeEHR}, Teoh et al. \cite{Teoh2018StrokeEHR} and Yang et al. \cite{Yang2021StrokePrediction}. The reference models excluded PPG-derived morphology, free-text features, imaging data and downstream treatment trajectories. We implemented these references using the structured predictors available in the study cohorts and the information domains described in the cited publications.

\par For warning analyses, structured-EHR variables were required to be available at or before the corresponding candidate evaluation time. For clinical-course analyses, EHR variables were restricted to the first 360\,min after ICU admission in MIMIC-III or emergency-department arrival in MC-MED, truncated at earlier ICU discharge or emergency-department departure. Eligible predictors comprised demographics, pre-existing comorbidities, laboratory measurements and vital signs available within the relevant interval. The available-component \(\mathrm{CHA}_2\mathrm{DS}_2\text{-}\mathrm{VASc}\) reference was defined as

\[
S_i^{\mathrm{CHA}}
=
2\mathbb{I}(A_i\geq75)
+
\mathbb{I}(65\leq A_i<75)
+
\mathbb{I}(F_i=1)
+
\mathbb{I}(H_i=1)
+
\mathbb{I}(D_i=1)
+
\mathbb{I}(V_i=1),
\]

where $A_i$ denotes age, $F_i$ female sex, $H_i$ hypertension, $D_i$ diabetes and $V_i$ ischaemic heart disease as the available vascular-disease component. This available-component score uses the cross-dataset variable map shown in the equation. Its deterministic rule and fixed threshold yield identical MC-MED estimates across five applications (s.d.\ 0.0000). For evaluation, it was normalised as

\[
\widetilde{S}_i^{\mathrm{CHA}}
=
\frac{S_i^{\mathrm{CHA}}}{6}
\in[0,1].
\]

\par Nwosu--EHR \cite{Nwosu2019StrokeEHR} used age, sex, hypertension, diabetes, ischaemic heart disease and the most recent glucose measurement. Teoh--EHR \cite{Teoh2018StrokeEHR} used age and sex; hypertension, diabetes, hyperlipidaemia, ischaemic heart disease and chronic kidney disease; the most recent creatinine, glucose, haematocrit, potassium and sodium measurements; and the mean, maximum and minimum diastolic blood pressure, heart rate, respiratory rate, systolic blood pressure and peripheral oxygen saturation. Yang--EHR \cite{Yang2021StrokePrediction} used age and sex; hypertension, diabetes, hyperlipidaemia, ischaemic heart disease and chronic kidney disease; the most recent glucose measurement; and mean systolic blood pressure, diastolic blood pressure and heart rate.

\par Nwosu--EHR used a Random Forest \cite{Breiman2001RandomForests}, whereas Teoh--EHR and Yang--EHR used XGBoost \cite{Chen2016XGBoost}. For the three learned EHR references, missing values were imputed using medians estimated from the corresponding MIMIC-III training partition, RobustScaler parameters were fitted on the same partition, and the Synthetic Minority Over-sampling Technique was restricted to training data \cite{Chawla2002SMOTE}. The available-component \(\mathrm{CHA}_2\mathrm{DS}_2\text{-}\mathrm{VASc}\) reference used training-partition median imputation for missing available components and no learned estimator. Preprocessing parameters, fitted EHR estimators and decision thresholds determined in MIMIC-III were then applied to MC-MED for frozen cross-dataset evaluation.

\subsection*{Model evaluation and statistical analysis}

\par Age is reported as mean (s.d.) or median [IQR], and categorical variables as $n$ (\%) with the available cohort denominator. Comorbidities are counted by recorded category. The SQI population comprises records evaluated before temporal and window-retention screening.

Strict internal warning performance used patient-grouped five-fold cross-validation; historical fold-wise metrics used recording groups with the patient overlap described in Results. Accuracy, precision, recall, \(F_1\), \(F_2\) and AUC were calculated at the window level. The patient was the resampling unit; labelled windows were the evaluation units. For clinical-course stratification, patient-grouped five-fold cross-validation was used, with hospitalisations from the same patient kept within one fold.

\par Fold-specific metrics summarised each fitted model's predictions as mean and s.d. Ensemble metrics averaged the five model probabilities and used patient-cluster bootstrap intervals. Patient-grouped analyses used training-fold preprocessing.

\par For the PPG warning model, epoch and decision-threshold selection followed the procedure described above. For the separately fitted structured-EHR clinical-course reference models \cite{Nwosu2019StrokeEHR,Teoh2018StrokeEHR,Yang2021StrokePrediction}, MIMIC-III development used five-fold stratified group cross-validation with patient identity as the grouping variable. Decision thresholds were selected using grouped out-of-fold predictions generated within each outer-training partition, with Youden's index used to define the operating threshold. This threshold-selection procedure was specific to the structured-EHR clinical-course references.

\par The primary reported threshold-dependent metric was \(F_1\); accuracy, precision and recall were secondary threshold-dependent metrics. We additionally reported \(F_2\),

\[
F_2=
\frac{5\,\mathrm{precision}\times\mathrm{recall}}
{4\,\mathrm{precision}+\mathrm{recall}},
\]

which assigns greater weight to recall; AUC uses continuous scores before thresholding. Original performance is reported as mean $\pm$ s.d. across five MIMIC-III folds or five corresponding MC-MED applications. Table~\ref{tab:prognosis-performance} reports $F_1$, precision and recall for the modality-specific clinical-course cohorts.

\par We calculated SHAP values after model development to characterise feature attribution \cite{Lundberg2020SHAP}. The attribution figure shows representative MIMIC-III predictions for warning-state classification and clinical-course stratification.

\par We examined clinical-risk category, reported race, age and sex in exploratory subgroup analyses of the frozen MC-MED PPG warning ensemble. The five fold-specific applications provided descriptive \(F_1\) means and standard errors of the mean (s.e.m.). Patient-cluster bootstrap resampling of the mean-probability ensemble provided 95\% confidence intervals. Subgroup comparisons used patient-level permutation testing with 10,000 permutations. For variables with more than two groups, the omnibus statistic was the maximum absolute pairwise \(F_1\) difference, with Bonferroni correction for pairwise comparisons. All statistical tests were two-sided with \(\alpha=0.05\).

\subsection*{False-alert on control patients}

\par False-alert burden was evaluated by applying the horizon-specific warning models to the high-risk non-stroke cohorts defined above. For dataset \(D\) and nominal setting \(h\in\{240,300,360\}\) min, the window-level false-positive rate was the proportion of retained non-stroke windows classified as positive. A persistent alert required at least five positive windows in uninterrupted temporal sequence; gaps or interrupted recordings broke the sequence. The control analysis unit was the patient in MIMIC-III and the MRN group in MC-MED. The control-level persistent-alert rate was the proportion of eligible analysis units satisfying this rule. Stroke-cohort true-positive rate was the mean window-level recall across the five model applications at the same decision thresholds. Under this protocol the control cohorts yielded a mean of 12.7 evaluable windows per control unit in MIMIC-III (2,910 windows across 230 units) and 7.8 per control unit in MC-MED (1,284 windows across 164 units) at the 360-min setting.

\par Descriptive 95\% Wilson score intervals were calculated for control-level persistent-alert proportions from the integer alerted and total control-unit counts. The binomial unit was the patient in MIMIC-III and the MRN group in MC-MED. Figure~\ref{fig:alert-burden}d shows these control-level intervals.

\par The warning cohorts and original threshold-based results extend our earlier report \cite{Liu2026InHospitalStroke}. The present study adds clinical-course stratification, patient-grouped evaluation, simultaneous physiological measurements and monitoring-burden analyses.

\subsection*{Threshold-independent discrimination and confidence intervals}

\par AUROC was computed from continuous probabilities. Ensemble estimates averaged five model probabilities. Ninety-five per cent confidence intervals used 2,000 patient-cluster bootstrap resamples, drawing patients or MRN groups with all their windows and admissions, and paired model contrasts used identical patient-grouped windows. Tests were two-sided with $\alpha=0.05$.

\subsection*{Heart-rate and heart-rate-variability comparators}

\par We extracted pulse HR, SDNN, RMSSD, pNN50 and LF/HF from the same 500-beat windows used by the patient-grouped morphology model. HR and HR+HRV comparators used $L_2$ logistic regression with training-fold median imputation, missing indicators, standardisation and class weights. Three patient-grouped inner folds selected $C\in\{0.01,0.1,1,10,100\}$; five seeded fits used the same outer assignment as the morphology model. Paired contrasts used all 15,117 windows from 164 patients and 2,000 patient-cluster bootstrap replicates.

\subsection*{Matched controls, alert burden and available-segment analysis}

\par Candidate controls were monitored patients without documented in-hospital stroke who met the waveform-availability and signal-quality criteria. A propensity score for in-hospital stroke was estimated by logistic regression on age, sex, hypertension, diabetes, atrial fibrillation, ischaemic heart disease, admission type (surgical versus medical in MIMIC-III; emergency-severity index in MC-MED) and monitored duration before the event or pseudo-anchor. We matched stroke patients 1:3 without replacement to nearest neighbours within a calliper of 0.2 s.d. of the logit. Standardised mean differences assessed covariate balance. Matched controls received pseudo-anchors sampled from the stroke-cohort distribution of elapsed monitoring time, with at least 480\,min of preceding monitoring, and supplied the comparison group for the synchronous crest-time analysis. Alert burden was computed as the number of persistent alerts (\(\geq k\) consecutive positive windows, \(k=1,\dots,10\)) among controls divided by total monitored control time, expressed per 100 patient-days, with patient-cluster bootstrap CIs.

\par The available-segment analysis used a fixed probability threshold of 0.791 and required five consecutive positive windows within each recording segment. Confirmation occurred at the fifth window end. For each patient, we selected the earliest confirmed relative time across records and summarised its interval to the corresponding recognition anchor among alerted patients. Coverage used all eligible patients, including those with fewer than five windows. The 312 MIMIC-III records represented 175 admissions, 69 of which contained more than one recorded recognition anchor. Replaying the rule with resets at nonconsecutive indices, gaps exceeding 60\,s or backward coordinates yielded the same alert summary.

\par Analysed duration was the union of absolute window intervals within each patient, summed over disjoint intervals across admissions and anchors. Patient-grouped windows used raw timestamps; the available-segment analysis used reconstructed recording coordinates.

\subsection*{Clinical subgroups and diagnostic strata}

\par Stroke-type definitions (ischaemic versus haemorrhagic) used adjudicated diagnoses and ICD codes. Pre-anchor diagnoses or rhythm annotations defined atrial fibrillation. Peri-procedural status required an anchor within 24\,h of cardiac or vascular surgery or an endovascular procedure, using a validated procedure timestamp when available. Vasoactive or sedative exposure came from medication-administration records within 6\,h before the warning interval. Subgroup $n$ counts unique patients within each stratum. AF strata partition the 164 strict MIMIC-III patients. Drug-exposure strata contain 60 exposed and 129 unexposed patients, with 25 contributing to both across distinct admission--anchor groups and preceding 6-h exposure windows. The union is 164 patients; classification is consistent within each admission--anchor group. MC-MED $F_1$ confidence intervals use 10,000 patient-cluster resamples of the separate historical ensemble subset. 
MIMIC-III subgroup $F_1$ intervals used 2,000 patient-cluster bootstrap samples at the fixed cutoff of 0.5; the original MC-MED subgroup $F_1$ comparisons used patient-level permutations.

\par Diagnostic strata retained the original Table~\ref{tab:cohort-characteristics} cohorts and their admission or visit codes, separating ischaemic codes, haemorrhagic codes, both code families and other codes. The 360-min warning analysis used the mean of the five stored model probabilities and a cutoff of 0.5. Each stratum reports patients and patient-cluster intervals; diagnostic-code groups describe recorded disease.

\subsection*{Synchronous crest-time and ECG measurements}
\par PPG--ECG measurements used the selected synchronous 20--30-s segments. PLETH underwent pyPPG fourth-order Chebyshev-II filtering at 0.5--12 Hz and 50-ms smoothing. Crest time was $1000(sp-on)/f_s$ in milliseconds, with at least ten valid paired beats per segment. ECG peaks used WFDB XQRS; HR required at least ten RR intervals and at least 90\% between 0.30 and 2.0 s. ECG-to-pulse HR ratios had to lie between 0.8 and 1.2. This yielded 60 stroke segments and 101 clinical non-stroke segments. Individual Pearson correlations required at least three variable observations. Within-patient regression included HR per 10 bpm and hours relative to the assigned anchor, with patient intercepts and equal total weight per patient. Temporal regression used patients observed in at least two time bins: 15 stroke patients with 55 segments and 20 controls with 100 segments. Percentile intervals used 2,000 patient-cluster bootstrap samples within each group. Hourly descriptive profiles use patient means within observed time bins; no observations were extrapolated beyond the recorded support.

\par To assess temporal variation in the association itself, we used the common $-8$ to $-2$\,h observation range. Repeated-measures correlations\cite{Bakdash2017Repeated} were calculated separately for the $-8$ to $-5$\,h and $-5$ to $-2$\,h periods, using patients with at least two observations and non-zero variance in both measurements within the corresponding period. Late-minus-early correlation differences retained each patient's observations together during 2,000 bootstrap resamples, stratified by group and early-only, late-only or both-period support. Early and late estimates included 5 and 7 stroke patients and 17 and 16 controls, respectively.

\par For the paired $F_1$ and clinical-stratum summaries, we used a common probability cutoff of 0.5, without selecting a cutoff from the evaluation labels. ResNet-1D probabilities and HR/HRV probabilities were matched by window identifier and patient fold. For each of 2,000 bootstrap draws, patients were sampled within their observed outcome-support stratum (event-only, non-event-only or both); all windows for a selected patient remained together. Paired differences used the same draw for both models. Hypertension and diabetes were linked from the existing admission diagnosis flags; AF and six-hour drug exposure used the existing clinical subgroup definitions.

\subsection*{Cohort-level feature attribution}

\par SHAP values were computed with the DeepExplainer for every evaluated window in both cohorts, using 500 background windows sampled from the MIMIC-III training folds. Cohort-level importance was the mean absolute SHAP value per feature; agreement between cohorts was quantified by Spearman's rank correlation across the 17 features and by the proportion of features with the same sign of the mean SHAP--feature-value correlation. Time-resolved attribution averaged absolute SHAP values within hourly bins before the documented-recognition anchor.

\subsection*{Signal quality and waveform retention}

\par Signal quality and waveform retention were characterised to assess the availability of recordings suitable for beat-level morphology extraction. Patient-level signal quality was summarised using the pyPPG SQI, for which higher values indicate greater morphological consistency for fiducial extraction \cite{Goda2024pyPPG}.

\par For warning-state classification, hourly waveform retention was defined as the proportion of eligible hourly intervals from which usable PPG remained after temporal alignment, signal-quality filtering and waveform processing. For clinical-course stratification, retention was additionally characterised at the DAT unit level as the proportion of eligible DAT units yielding usable pyPPG output; duration retention was the proportion of eligible recording time retained in the source-record processing analysis; and processing success was the proportion of attempted DAT units processed successfully. SQI, retention, recording-duration and processing-success measures were not included among model predictors.

\FloatBarrier

\begingroup\fontsize{8.5}{9.5}\selectfont

\endgroup

\begingroup\fontsize{9}{10}\selectfont
\titlespacing*{\section}{0pt}{5pt}{2pt}
\section*{Data availability}
MIMIC-III is available through PhysioNet under credentialed access and the applicable data-use agreement (\url{https://doi.org/10.13026/C2XW26}). MC-MED version 1.0.0 is available through PhysioNet under credentialed access, required training and the applicable data-use agreement (\url{https://doi.org/10.13026/jz99-4j81}). The derived aggregate tables needed to interpret the reported results are included in the article. Patient-level derived features, waveform-linked intermediate files and restricted source data are not publicly redistributed because the applicable source-dataset agreements restrict redistribution. Researchers seeking access to the source data must apply through PhysioNet and satisfy the relevant credentialing, training, institutional and data-use requirements.

\section*{Code availability}
Analysis code is available at \url{https://github.com/6jm233333/ppg-stroke-continuous-management}. Additional scripts, configurations and aggregate source data accompany the revised manuscript.

\section*{Acknowledgements}
This work was supported by the National Natural Science Foundation of China (Grant No. 32541017), the Opening Foundation of the State Key Laboratory of Transvascular Implantation Devices (Grant No. SKLTID2025102), the Fundamental Research Funds for the Central Universities (Grant No. N22025028), and the Jiangsu Funding Program for Excellent Postdoctoral Talent.

\section*{Author contributions}
C.D. and J.L. contributed equally to study conception, experimental design, data analysis and manuscript preparation. J.W. and H.X. reviewed the manuscript and contributed to rigorous verification of the dataset labels. J.W. and D.Z. jointly supervised the study and serve as corresponding authors.

\section*{Competing interests}
Dr. Cheng Ding, the first author of the manuscript, serves as one of the editorial team in this collection.

\endgroup

\begin{thebibliography}{10}
\expandafter\ifx\csname url\endcsname\relax
  \def\url#1{\texttt{#1}}\fi
\expandafter\ifx\csname urlprefix\endcsname\relax\def\urlprefix{URL }\fi
\expandafter\ifx\csname doiprefix\endcsname\relax\def\doiprefix{DOI }\fi
\providecommand{\bibinfo}[2]{#2}
\providecommand{\eprint}[2][]{\url{#2}}

\bibitem{Feigin2021GBD2019}
\bibinfo{author}{{GBD 2019 Stroke Collaborators}}.
\newblock \bibinfo{journal}{\bibinfo{title}{Global, regional, and national
  burden of stroke and its risk factors, 1990--2019: a systematic analysis for
  the global burden of disease study 2019}}.
\newblock {\emph{\JournalTitle{The Lancet Neurology}}}
  \textbf{\bibinfo{volume}{20}}, \bibinfo{pages}{795--820}
  (\bibinfo{year}{2021}).
\newblock \doiprefix 10.1016/S1474-4422(21)00252-0.

\bibitem{Rothwell2010BPV}
\bibinfo{author}{Rothwell, P.~M.} \emph{et~al.}
\newblock \bibinfo{journal}{\bibinfo{title}{Prognostic significance of
  visit-to-visit variability, maximum systolic blood pressure, and episodic
  hypertension}}.
\newblock {\emph{\JournalTitle{The Lancet}}} \textbf{\bibinfo{volume}{375}},
  \bibinfo{pages}{895--905} (\bibinfo{year}{2010}).
\newblock \doiprefix 10.1016/S0140-6736(10)60308-X.

\bibitem{Laurent2003Stiffness}
\bibinfo{author}{Laurent, S.} \emph{et~al.}
\newblock \bibinfo{journal}{\bibinfo{title}{Aortic stiffness is an independent
  predictor of fatal stroke in essential hypertension}}.
\newblock {\emph{\JournalTitle{Stroke}}} \textbf{\bibinfo{volume}{34}},
  \bibinfo{pages}{1203--1206} (\bibinfo{year}{2003}).
\newblock \doiprefix 10.1161/01.STR.0000065428.03209.64.

\bibitem{GBD2021StrokeRiskFactorCollaborators2024}
\bibinfo{author}{{GBD 2021 Stroke Risk Factor Collaborators}}.
\newblock \bibinfo{journal}{\bibinfo{title}{Global, regional, and national
  burden of stroke and its risk factors, 1990--2021: a systematic analysis for
  the global burden of disease study 2021}}.
\newblock {\emph{\JournalTitle{The Lancet Neurology}}}
  \textbf{\bibinfo{volume}{23}}, \bibinfo{pages}{973--1003}
  (\bibinfo{year}{2024}).
\newblock \doiprefix 10.1016/S1474-4422(24)00369-7.

\bibitem{Lip2010CHA2DS2VASc}
\bibinfo{author}{Lip, G. Y.~H.}, \bibinfo{author}{Nieuwlaat, R.},
  \bibinfo{author}{Pisters, R.}, \bibinfo{author}{Lane, D.~A.} \&
  \bibinfo{author}{Crijns, H. J. G.~M.}
\newblock \bibinfo{journal}{\bibinfo{title}{Refining clinical risk
  stratification for predicting stroke and thromboembolism in atrial
  fibrillation using a novel risk factor-based approach: the {Euro Heart
  Survey} on atrial fibrillation}}.
\newblock {\emph{\JournalTitle{Chest}}} \textbf{\bibinfo{volume}{137}},
  \bibinfo{pages}{263--272} (\bibinfo{year}{2010}).
\newblock \doiprefix 10.1378/chest.09-1584.

\bibitem{Kim2017CHA2DS2VASc}
\bibinfo{author}{Kim, T.-H.} \emph{et~al.}
\newblock \bibinfo{journal}{\bibinfo{title}{{{CHA$_2$DS$_2$-VASc}} score
  (congestive heart failure, hypertension, age $\geq 75$ [doubled], diabetes
  mellitus, prior stroke or transient ischemic attack [doubled], vascular
  disease, age 65--74, female) for stroke in asian patients with atrial
  fibrillation: a korean nationwide sample cohort study}}.
\newblock {\emph{\JournalTitle{Stroke}}} \textbf{\bibinfo{volume}{48}},
  \bibinfo{pages}{1524--1530} (\bibinfo{year}{2017}).
\newblock \doiprefix 10.1161/STROKEAHA.117.016926.

\bibitem{Nwosu2019StrokeEHR}
\bibinfo{author}{Nwosu, C.~S.}, \bibinfo{author}{Dev, S.},
  \bibinfo{author}{Bhardwaj, P.}, \bibinfo{author}{Veeravalli, B.} \&
  \bibinfo{author}{John, D.}
\newblock \bibinfo{journal}{\bibinfo{title}{Predicting stroke from electronic
  health records}}.
\newblock {\emph{\JournalTitle{2019 41st Annual International Conference of the
  {IEEE} Engineering in Medicine and Biology Society ({EMBC})}}}
  \bibinfo{pages}{5704--5707} (\bibinfo{year}{2019}).
\newblock \doiprefix 10.1109/EMBC.2019.8857234.

\bibitem{Teoh2018StrokeEHR}
\bibinfo{author}{Teoh, D.}
\newblock \bibinfo{journal}{\bibinfo{title}{Towards stroke prediction using
  electronic health records}}.
\newblock {\emph{\JournalTitle{BMC Medical Informatics and Decision Making}}}
  \textbf{\bibinfo{volume}{18}}, \bibinfo{pages}{127} (\bibinfo{year}{2018}).
\newblock \doiprefix 10.1186/s12911-018-0702-y.

\bibitem{Yang2021StrokePrediction}
\bibinfo{author}{Yang, Y.}, \bibinfo{author}{Zheng, J.}, \bibinfo{author}{Du,
  Z.}, \bibinfo{author}{Li, Y.} \& \bibinfo{author}{Cai, Y.}
\newblock \bibinfo{journal}{\bibinfo{title}{Accurate prediction of stroke for
  hypertensive patients based on medical big data and machine learning
  algorithms: retrospective study}}.
\newblock {\emph{\JournalTitle{JMIR Medical Informatics}}}
  \textbf{\bibinfo{volume}{9}}, \bibinfo{pages}{e30277} (\bibinfo{year}{2021}).
\newblock \doiprefix 10.2196/30277.

\bibitem{Smyth2022INTERSTROKE}
\bibinfo{author}{Smyth, A.} \emph{et~al.}
\newblock \bibinfo{journal}{\bibinfo{title}{Anger or emotional upset and heavy
  physical exertion as triggers of stroke: the {INTERSTROKE} study}}.
\newblock {\emph{\JournalTitle{European Heart Journal}}}
  \textbf{\bibinfo{volume}{43}}, \bibinfo{pages}{202} (\bibinfo{year}{2022}).

\bibitem{Noda2023ICUStroke}
\bibinfo{author}{Noda, K.}, \bibinfo{author}{Koga, M.} \&
  \bibinfo{author}{Toyoda, K.}
\newblock \bibinfo{journal}{\bibinfo{title}{Recognition of strokes in the
  {ICU}: a narrative review}}.
\newblock {\emph{\JournalTitle{Journal of Cardiovascular Development and
  Disease}}} \textbf{\bibinfo{volume}{10}}, \bibinfo{pages}{182}
  (\bibinfo{year}{2023}).
\newblock \doiprefix 10.3390/jcdd10040182.

\bibitem{Qureshi2008AHR}
\bibinfo{author}{Qureshi, A.~I.}
\newblock \bibinfo{journal}{\bibinfo{title}{Acute hypertensive response in
  patients with stroke: pathophysiology and management}}.
\newblock {\emph{\JournalTitle{Circulation}}} \textbf{\bibinfo{volume}{118}},
  \bibinfo{pages}{176--187} (\bibinfo{year}{2008}).
\newblock \doiprefix 10.1161/CIRCULATIONAHA.107.723874.

\bibitem{Scheitz2022StrokeHeart}
\bibinfo{author}{Scheitz, J.~F.} \emph{et~al.}
\newblock \bibinfo{journal}{\bibinfo{title}{Stroke--heart syndrome: recent
  advances and challenges}}.
\newblock {\emph{\JournalTitle{Journal of the American Heart Association}}}
  \textbf{\bibinfo{volume}{11}}, \bibinfo{pages}{e026528}
  (\bibinfo{year}{2022}).
\newblock \doiprefix 10.1161/JAHA.122.026528.

\bibitem{Intharakham2019Autoregulation}
\bibinfo{author}{Intharakham, K.}, \bibinfo{author}{Beishon, L.},
  \bibinfo{author}{Panerai, R.~B.}, \bibinfo{author}{Haunton, V.~J.} \&
  \bibinfo{author}{Robinson, T.~G.}
\newblock \bibinfo{journal}{\bibinfo{title}{Assessment of cerebral
  autoregulation in stroke: a systematic review and meta-analysis of studies at
  rest}}.
\newblock {\emph{\JournalTitle{Journal of Cerebral Blood Flow \& Metabolism}}}
  \textbf{\bibinfo{volume}{39}}, \bibinfo{pages}{2105--2116}
  (\bibinfo{year}{2019}).
\newblock \doiprefix 10.1177/0271678X19871013.

\bibitem{Charlton2022WearablePPG}
\bibinfo{author}{Charlton, P.~H.} \emph{et~al.}
\newblock \bibinfo{journal}{\bibinfo{title}{Wearable photoplethysmography for
  cardiovascular monitoring}}.
\newblock {\emph{\JournalTitle{Proceedings of the IEEE}}}
  \textbf{\bibinfo{volume}{110}}, \bibinfo{pages}{355--381}
  (\bibinfo{year}{2022}).
\newblock \doiprefix 10.1109/JPROC.2022.3149785.

\bibitem{Charlton2023WearablePPGRoadmap}
\bibinfo{author}{Charlton, P.~H.}, \bibinfo{author}{Allen, J.},
  \bibinfo{author}{Bail{\'o}n, R.} \emph{et~al.}
\newblock \bibinfo{journal}{\bibinfo{title}{The 2023 wearable
  photoplethysmography roadmap}}.
\newblock {\emph{\JournalTitle{Physiological Measurement}}}
  \textbf{\bibinfo{volume}{44}}, \bibinfo{pages}{111001}
  (\bibinfo{year}{2023}).
\newblock \doiprefix 10.1088/1361-6579/acead2.

\bibitem{Allen2007PPG}
\bibinfo{author}{Allen, J.}
\newblock \bibinfo{journal}{\bibinfo{title}{Photoplethysmography and its
  application in clinical physiological measurement}}.
\newblock {\emph{\JournalTitle{Physiological Measurement}}}
  \textbf{\bibinfo{volume}{28}}, \bibinfo{pages}{R1--R39}
  (\bibinfo{year}{2007}).
\newblock \doiprefix 10.1088/0967-3334/28/3/R01.

\bibitem{Charlton2022Hemodynamics}
\bibinfo{author}{Charlton, P.~H.} \emph{et~al.}
\newblock \bibinfo{journal}{\bibinfo{title}{Assessing hemodynamics from the
  photoplethysmogram to gain insights into vascular age: a review from
  {VascAgeNet}}}.
\newblock {\emph{\JournalTitle{American Journal of Physiology-Heart and
  Circulatory Physiology}}} \textbf{\bibinfo{volume}{322}},
  \bibinfo{pages}{H493--H522} (\bibinfo{year}{2022}).
\newblock \doiprefix 10.1152/ajpheart.00392.2021.

\bibitem{Takazawa1998APG}
\bibinfo{author}{Takazawa, K.}, \bibinfo{author}{Tanaka, N.},
  \bibinfo{author}{Fujita, M.} \emph{et~al.}
\newblock \bibinfo{journal}{\bibinfo{title}{Assessment of vasoactive agents and
  vascular aging by the second derivative of photoplethysmogram waveform}}.
\newblock {\emph{\JournalTitle{Hypertension}}} \textbf{\bibinfo{volume}{32}},
  \bibinfo{pages}{365--370} (\bibinfo{year}{1998}).
\newblock \doiprefix 10.1161/01.HYP.32.2.365.

\bibitem{Elgendi2024PPGBloodPressure}
\bibinfo{author}{Elgendi, M.} \emph{et~al.}
\newblock \bibinfo{journal}{\bibinfo{title}{Recommendations for evaluating
  photoplethysmography-based algorithms for blood pressure assessment}}.
\newblock {\emph{\JournalTitle{Communications Medicine}}}
  \textbf{\bibinfo{volume}{4}}, \bibinfo{pages}{140} (\bibinfo{year}{2024}).
\newblock \doiprefix 10.1038/s43856-024-00555-2.

\bibitem{Ding2026CardioPPG}
\bibinfo{author}{Ding, Z.} \emph{et~al.}
\newblock \bibinfo{journal}{\bibinfo{title}{{AI} modeling photoplethysmography
  to electrocardiography useful for predicting cardiovascular disease}}.
\newblock {\emph{\JournalTitle{npj Digital Medicine}}}
  \textbf{\bibinfo{volume}{9}}, \bibinfo{pages}{61} (\bibinfo{year}{2026}).
\newblock \doiprefix 10.1038/s41746-025-02228-3.

\bibitem{Perez2019AppleHeart}
\bibinfo{author}{Perez, M.~V.} \emph{et~al.}
\newblock \bibinfo{journal}{\bibinfo{title}{Large-scale assessment of a
  smartwatch to identify atrial fibrillation}}.
\newblock {\emph{\JournalTitle{New England Journal of Medicine}}}
  \textbf{\bibinfo{volume}{381}}, \bibinfo{pages}{1909--1917}
  (\bibinfo{year}{2019}).
\newblock \doiprefix 10.1056/NEJMoa1901183.

\bibitem{Guo2019HuaweiHeart}
\bibinfo{author}{Guo, Y.} \emph{et~al.}
\newblock \bibinfo{journal}{\bibinfo{title}{Mobile photoplethysmographic
  technology to detect atrial fibrillation}}.
\newblock {\emph{\JournalTitle{Journal of the American College of Cardiology}}}
  \textbf{\bibinfo{volume}{74}}, \bibinfo{pages}{2365--2375}
  (\bibinfo{year}{2019}).
\newblock \doiprefix 10.1016/j.jacc.2019.08.019.

\bibitem{Cumbler2014InHospital}
\bibinfo{author}{Cumbler, E.} \emph{et~al.}
\newblock \bibinfo{journal}{\bibinfo{title}{Quality of care and outcomes for
  in-hospital ischemic stroke: findings from the national {Get With The
  Guidelines-Stroke}}}.
\newblock {\emph{\JournalTitle{Stroke}}} \textbf{\bibinfo{volume}{45}},
  \bibinfo{pages}{231--238} (\bibinfo{year}{2014}).
\newblock \doiprefix 10.1161/STROKEAHA.113.003617.

\bibitem{Akbik2020InHospital}
\bibinfo{author}{Akbik, F.} \emph{et~al.}
\newblock \bibinfo{journal}{\bibinfo{title}{Trends in reperfusion therapy for
  in-hospital ischemic stroke in the endovascular therapy era}}.
\newblock {\emph{\JournalTitle{JAMA Neurology}}} \textbf{\bibinfo{volume}{77}},
  \bibinfo{pages}{1486--1495} (\bibinfo{year}{2020}).
\newblock \doiprefix 10.1001/jamaneurol.2020.3362.

\bibitem{Correia2025DELAY}
\bibinfo{author}{Correia, P.~N.} \emph{et~al.}
\newblock \bibinfo{journal}{\bibinfo{title}{Extended delays in recognition of
  stroke symptoms and stroke code activation for in-hospital strokes: the
  {DELAY} study}}.
\newblock {\emph{\JournalTitle{Canadian Journal of Neurological Sciences}}}
  \textbf{\bibinfo{volume}{52}}, \bibinfo{pages}{502--509}
  (\bibinfo{year}{2025}).
\newblock \doiprefix 10.1017/cjn.2024.304.

\bibitem{Johnson2016MIMICIII}
\bibinfo{author}{Johnson, A. E.~W.} \emph{et~al.}
\newblock \bibinfo{journal}{\bibinfo{title}{{MIMIC-III}, a freely accessible
  critical care database}}.
\newblock {\emph{\JournalTitle{Scientific Data}}} \textbf{\bibinfo{volume}{3}},
  \bibinfo{pages}{160035} (\bibinfo{year}{2016}).
\newblock \doiprefix 10.1038/sdata.2016.35.

\bibitem{Moody2020MIMICIIIWaveform}
\bibinfo{author}{Moody, B.}, \bibinfo{author}{Moody, G.},
  \bibinfo{author}{Villarroel, M.}, \bibinfo{author}{Clifford, G.~D.} \&
  \bibinfo{author}{Silva, I.}
\newblock \bibinfo{journal}{\bibinfo{title}{{MIMIC-III Waveform Database
  Matched Subset}}}.
\newblock {\emph{\JournalTitle{PhysioNet}}}  (\bibinfo{year}{2020}).
\newblock \urlprefix\url{https://doi.org/10.13026/c2294b}.
\newblock \doiprefix 10.13026/c2294b.
\newblock \bibinfo{note}{Version 1.0}.

\bibitem{Kansal2025MCMED}
\bibinfo{author}{Kansal, A.}, \bibinfo{author}{Chen, E.}, \bibinfo{author}{Jin,
  B.~T.}, \bibinfo{author}{Rajpurkar, P.} \& \bibinfo{author}{Kim, D.~A.}
\newblock \bibinfo{journal}{\bibinfo{title}{{MC-MED}, multimodal clinical
  monitoring in the emergency department}}.
\newblock {\emph{\JournalTitle{Scientific Data}}}
  \textbf{\bibinfo{volume}{12}}, \bibinfo{pages}{1094} (\bibinfo{year}{2025}).
\newblock \urlprefix\url{https://doi.org/10.1038/s41597-025-05419-5}.
\newblock \doiprefix 10.1038/s41597-025-05419-5.

\bibitem{Kansal2025MCMEDDataset}
\bibinfo{author}{Kansal, A.}, \bibinfo{author}{Chen, E.}, \bibinfo{author}{Jin,
  T.}, \bibinfo{author}{Rajpurkar, P.} \& \bibinfo{author}{Kim, D.}
\newblock \bibinfo{journal}{\bibinfo{title}{{Multimodal Clinical Monitoring in
  the Emergency Department (MC-MED)}}}.
\newblock {\emph{\JournalTitle{PhysioNet}}}  (\bibinfo{year}{2025}).
\newblock \urlprefix\url{https://doi.org/10.13026/jz99-4j81}.
\newblock \doiprefix 10.13026/jz99-4j81.
\newblock \bibinfo{note}{Version 1.0.0}.

\bibitem{Liu2026InHospitalStroke}
\bibinfo{author}{Liu, J.}, \bibinfo{author}{Ding, C.}, \bibinfo{author}{Wu,
  J.}, \bibinfo{author}{Xu, H.} \& \bibinfo{author}{Zhang, D.}
\newblock \bibinfo{title}{In-hospital stroke risk-state classification from
  ppg-derived hemodynamic features} (\bibinfo{year}{2026}).
\newblock \doiprefix 10.48550/arXiv.2602.09328.
\newblock \eprint{2602.09328v2}.

\bibitem{Manning2014BPV}
\bibinfo{author}{Manning, L.}, \bibinfo{author}{Hirakawa, Y.},
  \bibinfo{author}{Arima, H.} \emph{et~al.}
\newblock \bibinfo{journal}{\bibinfo{title}{Blood pressure variability and
  outcome after acute intracerebral haemorrhage: a post-hoc analysis of
  {INTERACT2}, a randomised controlled trial}}.
\newblock {\emph{\JournalTitle{The Lancet Neurology}}}
  \textbf{\bibinfo{volume}{13}}, \bibinfo{pages}{364--373}
  (\bibinfo{year}{2014}).
\newblock \doiprefix 10.1016/S1474-4422(14)70018-3.

\bibitem{Macdonald2014DCI}
\bibinfo{author}{Macdonald, R.~L.}
\newblock \bibinfo{journal}{\bibinfo{title}{Delayed neurological deterioration
  after subarachnoid haemorrhage}}.
\newblock {\emph{\JournalTitle{Nature Reviews Neurology}}}
  \textbf{\bibinfo{volume}{10}}, \bibinfo{pages}{44--58}
  (\bibinfo{year}{2014}).
\newblock \doiprefix 10.1038/nrneurol.2013.246.

\bibitem{Tarnutzer2017EDMisdiagnosis}
\bibinfo{author}{Tarnutzer, A.~A.}, \bibinfo{author}{Lee, S.-H.},
  \bibinfo{author}{Robinson, K.~A.} \emph{et~al.}
\newblock \bibinfo{journal}{\bibinfo{title}{{ED} misdiagnosis of
  cerebrovascular events in the era of modern neuroimaging: a meta-analysis}}.
\newblock {\emph{\JournalTitle{Neurology}}} \textbf{\bibinfo{volume}{88}},
  \bibinfo{pages}{1468--1477} (\bibinfo{year}{2017}).
\newblock \doiprefix 10.1212/WNL.0000000000003814.

\bibitem{Drew2014AlarmFatigue}
\bibinfo{author}{Drew, B.~J.} \emph{et~al.}
\newblock \bibinfo{journal}{\bibinfo{title}{Insights into the problem of alarm
  fatigue with physiologic monitor devices: a comprehensive observational study
  of consecutive intensive care unit patients}}.
\newblock {\emph{\JournalTitle{PLoS ONE}}} \textbf{\bibinfo{volume}{9}},
  \bibinfo{pages}{e110274} (\bibinfo{year}{2014}).
\newblock \doiprefix 10.1371/journal.pone.0110274.

\bibitem{Huang2023VCPM}
\bibinfo{author}{Huang, B.} \emph{et~al.}
\newblock \bibinfo{journal}{\bibinfo{title}{Challenges and prospects of visual
  contactless physiological monitoring in clinical study}}.
\newblock {\emph{\JournalTitle{npj Digital Medicine}}}
  \textbf{\bibinfo{volume}{6}}, \bibinfo{pages}{231} (\bibinfo{year}{2023}).
\newblock \doiprefix 10.1038/s41746-023-00973-x.

\bibitem{Sjoding2020PulseOx}
\bibinfo{author}{Sjoding, M.~W.}, \bibinfo{author}{Dickson, R.~P.},
  \bibinfo{author}{Iwashyna, T.~J.}, \bibinfo{author}{Gay, S.~E.} \&
  \bibinfo{author}{Valley, T.~S.}
\newblock \bibinfo{journal}{\bibinfo{title}{Racial bias in pulse oximetry
  measurement}}.
\newblock {\emph{\JournalTitle{New England Journal of Medicine}}}
  \textbf{\bibinfo{volume}{383}}, \bibinfo{pages}{2477--2478}
  (\bibinfo{year}{2020}).
\newblock \doiprefix 10.1056/NEJMc2029240.

\bibitem{Collins2024TRIPODAI}
\bibinfo{author}{Collins, G.~S.} \emph{et~al.}
\newblock \bibinfo{journal}{\bibinfo{title}{Tripod+ai statement: updated
  guidance for reporting clinical prediction models that use regression or
  machine learning methods}}.
\newblock {\emph{\JournalTitle{BMJ}}} \textbf{\bibinfo{volume}{385}},
  \bibinfo{pages}{e078378} (\bibinfo{year}{2024}).
\newblock \doiprefix 10.1136/bmj-2023-078378.

\bibitem{Norgeot2020MICLAIM}
\bibinfo{author}{Norgeot, B.} \emph{et~al.}
\newblock \bibinfo{journal}{\bibinfo{title}{Minimum information about clinical
  artificial intelligence modeling: the mi-claim checklist}}.
\newblock {\emph{\JournalTitle{Nature Medicine}}}
  \textbf{\bibinfo{volume}{26}}, \bibinfo{pages}{1320--1324}
  (\bibinfo{year}{2020}).
\newblock \doiprefix 10.1038/s41591-020-1041-y.

\bibitem{Yang2022EHRLLM}
\bibinfo{author}{Yang, X.} \emph{et~al.}
\newblock \bibinfo{journal}{\bibinfo{title}{A large language model for
  electronic health records}}.
\newblock {\emph{\JournalTitle{npj Digital Medicine}}}
  \textbf{\bibinfo{volume}{5}}, \bibinfo{pages}{194} (\bibinfo{year}{2022}).
\newblock \doiprefix 10.1038/s41746-022-00742-2.

\bibitem{Agrawal2022ClinicalLLM}
\bibinfo{author}{Agrawal, M.}, \bibinfo{author}{Hegselmann, S.},
  \bibinfo{author}{Lang, H.}, \bibinfo{author}{Kim, Y.} \&
  \bibinfo{author}{Sontag, D.}
\newblock \bibinfo{title}{Large language models are few-shot clinical
  information extractors}.
\newblock In \bibinfo{editor}{Goldberg, Y.}, \bibinfo{editor}{Kozareva, Z.} \&
  \bibinfo{editor}{Zhang, Y.} (eds.) \emph{\bibinfo{booktitle}{Proceedings of
  the 2022 Conference on Empirical Methods in Natural Language Processing}},
  \bibinfo{pages}{1998--2022} (\bibinfo{publisher}{Association for
  Computational Linguistics}, \bibinfo{address}{Abu Dhabi, United Arab
  Emirates}, \bibinfo{year}{2022}).
\newblock \doiprefix 10.18653/v1/2022.emnlp-main.130.

\bibitem{Goda2024pyPPG}
\bibinfo{author}{Goda, M.~{\'A}.}, \bibinfo{author}{Charlton, P.~H.} \&
  \bibinfo{author}{Behar, J.~A.}
\newblock \bibinfo{journal}{\bibinfo{title}{pyppg: a python toolbox for
  comprehensive photoplethysmography signal analysis}}.
\newblock {\emph{\JournalTitle{Physiological Measurement}}}
  \textbf{\bibinfo{volume}{45}}, \bibinfo{pages}{045001}
  (\bibinfo{year}{2024}).
\newblock \doiprefix 10.1088/1361-6579/ad33a2.

\bibitem{He2016ResNet}
\bibinfo{author}{He, K.}, \bibinfo{author}{Zhang, X.}, \bibinfo{author}{Ren,
  S.} \& \bibinfo{author}{Sun, J.}
\newblock \bibinfo{title}{Deep residual learning for image recognition}.
\newblock In \bibinfo{editor}{Agapito, L.}, \bibinfo{editor}{Berg, T.},
  \bibinfo{editor}{Kosecka, J.} \& \bibinfo{editor}{Zelnik-Manor, L.} (eds.)
  \emph{\bibinfo{booktitle}{Proceedings of the IEEE Conference on Computer
  Vision and Pattern Recognition}}, \bibinfo{pages}{770--778}
  (\bibinfo{publisher}{IEEE}, \bibinfo{address}{Las Vegas, NV, USA},
  \bibinfo{year}{2016}).
\newblock \doiprefix 10.1109/CVPR.2016.90.

\bibitem{Breiman2001RandomForests}
\bibinfo{author}{Breiman, L.}
\newblock \bibinfo{journal}{\bibinfo{title}{Random forests}}.
\newblock {\emph{\JournalTitle{Machine Learning}}}
  \textbf{\bibinfo{volume}{45}}, \bibinfo{pages}{5--32} (\bibinfo{year}{2001}).
\newblock \doiprefix 10.1023/A:1010933404324.

\bibitem{Chen2016XGBoost}
\bibinfo{author}{Chen, T.} \& \bibinfo{author}{Guestrin, C.}
\newblock \bibinfo{title}{{XGBoost}: A scalable tree boosting system}.
\newblock In \bibinfo{editor}{Krishnapuram, B.} \emph{et~al.} (eds.)
  \emph{\bibinfo{booktitle}{Proceedings of the 22nd ACM SIGKDD International
  Conference on Knowledge Discovery and Data Mining}},
  \bibinfo{pages}{785--794} (\bibinfo{publisher}{Association for Computing
  Machinery}, \bibinfo{address}{San Francisco, CA, USA}, \bibinfo{year}{2016}).
\newblock \doiprefix 10.1145/2939672.2939785.

\bibitem{Chawla2002SMOTE}
\bibinfo{author}{Chawla, N.~V.}, \bibinfo{author}{Bowyer, K.~W.},
  \bibinfo{author}{Hall, L.~O.} \& \bibinfo{author}{Kegelmeyer, W.~P.}
\newblock \bibinfo{journal}{\bibinfo{title}{Smote: Synthetic minority
  over-sampling technique}}.
\newblock {\emph{\JournalTitle{Journal of Artificial Intelligence Research}}}
  \textbf{\bibinfo{volume}{16}}, \bibinfo{pages}{321--357}
  (\bibinfo{year}{2002}).
\newblock \doiprefix 10.1613/jair.953.

\bibitem{Lundberg2020SHAP}
\bibinfo{author}{Lundberg, S.~M.} \emph{et~al.}
\newblock \bibinfo{journal}{\bibinfo{title}{From local explanations to global
  understanding with explainable ai for trees}}.
\newblock {\emph{\JournalTitle{Nature Machine Intelligence}}}
  \textbf{\bibinfo{volume}{2}}, \bibinfo{pages}{56--67} (\bibinfo{year}{2020}).
\newblock \doiprefix 10.1038/s42256-019-0138-9.

\bibitem{Bakdash2017Repeated}
\bibinfo{author}{Bakdash, J.~Z.} \& \bibinfo{author}{Marusich, L.~R.}
\newblock \bibinfo{journal}{\bibinfo{title}{Repeated measures correlation}}.
\newblock {\emph{\JournalTitle{Frontiers in Psychology}}}
  \textbf{\bibinfo{volume}{8}}, \bibinfo{pages}{456} (\bibinfo{year}{2017}).
\newblock \doiprefix 10.3389/fpsyg.2017.00456.

\end{thebibliography}
\end{document}